\documentclass[10pt,journal,compsoc]{IEEEtran}
\ifCLASSOPTIONcompsoc
  \usepackage[nocompress]{cite}
\else
  \usepackage{cite}
\fi

\ifCLASSINFOpdf
\usepackage[pdftex]{graphicx}
\graphicspath{{figures/}}
\else
\usepackage[dvips]{graphicx}
\graphicspath{{figures/}}
\fi
\usepackage{amsmath}
\usepackage{amsfonts}
\usepackage[ruled,norelsize]{algorithm2e}
\usepackage{wrapfig}
\usepackage[hidelinks]{hyperref}
\hypersetup{
    colorlinks=false,
    linkcolor=blue,
    filecolor=magenta,      
    urlcolor=cyan,
    pdftitle={MSFormer},
}
\usepackage{caption}

\usepackage{cite}                      % needed to automatically sort the references
\usepackage{tabu}                      % only used for the table example
\usepackage{booktabs}                  % only used for the table example
\usepackage{multirow} 
\usepackage{colortbl}
\usepackage{bigdelim}
\usepackage{fancyhdr}
\usepackage{enumitem}
\usepackage{setspace}
\usepackage{color}
\definecolor{myred}{RGB}{0,0,0}
\usepackage{subfiles}
\usepackage{amsmath}
\usepackage{mathtools}
\usepackage{float}
\usepackage{pgfplots} 
\usepackage{subcaption}
\usepackage{array}
\usepackage{tikz}

\newcolumntype{L}[1]{>{\raggedright\let\newline\\\arraybackslash\hspace{0pt}}m{#1}}
\newcolumntype{C}[1]{>{\centering\let\newline\\\arraybackslash\hspace{0pt}}m{#1}}
\newcolumntype{R}[1]{>{\raggedleft\let\newline\\\arraybackslash\hspace{0pt}}m{#1}}

\newcommand{\Section}[1]{\vspace{-2mm} \section{#1} \vspace{-1mm}}
\newcommand{\SubSection}[1]{\vspace{-1mm} \subsection{#1} \vspace{-1mm}}
\newcommand{\SubSubSection}[1]{\vspace{-1mm} \subsubsection{#1} \vspace{-1mm}}

\newcommand{\redrect}{\tikz[baseline=0ex]\draw[red, thick] (0,0) rectangle (0.2cm,0.2cm);}

\begin{document}


\title{ MSFormer: A Skeleton-multiview Fusion Method For Tooth Instance Segmentation
}
% \IEEEaftertitletext{\vspace{15em}}

\author{Yuan Li,
        Huan Liu,
        Yubo Tao*,
        Xiangyang He, \\
        Haifeng Li,
        Xiaohu Guo,
        Hai Lin*
\IEEEcompsocitemizethanks{
  \IEEEcompsocthanksitem
  Yuan Li, Huan Liu, Yubo Tao and Hai Lin are with Zhejiang University in China. Email: yuanli@zju.edu.cn, alisalh@zju.edu.cn, taoyubo@cad.zju.edu.cn, lin@zju.edu.cn.
  \IEEEcompsocthanksitem
  Haifeng Li is with the Central South University in China. Email: lihaifeng@csu.edu.cn.
  \IEEEcompsocthanksitem
  Xiaohu Guo is with the University of Texas at dallas.  Email: xguo@utdallas.edu.
}% <-this % stops an unwanted space
% \thanks{}
}

\IEEEtitleabstractindextext{%
\begin{abstract}
  Recently, deep learning–based tooth segmentation methods have been limited by the expensive and time-consuming processes of data collection and labeling. Achieving high-precision segmentation with limited datasets is critical. A viable solution to this entails fine-tuning pre-trained multiview-based models, thereby enhancing performance with limited data. However, relying solely on two-dimensional (2D) images for three-dimensional (3D) tooth segmentation can produce suboptimal outcomes because of occlusion and deformation, i.e., incomplete and distorted shape perception.
  To improve this fine-tuning-based solution, this paper advocates 2D–3D joint perception. The fundamental challenge in employing 2D–3D joint perception with limited data is that the 3D-related inputs and modules must follow a lightweight policy instead of using huge 3D data and parameter-rich modules that require extensive training data. Following this lightweight policy, this paper selects skeletons as the 3D inputs and introduces MSFormer, a novel method for tooth segmentation. MSFormer incorporates two lightweight modules into existing multiview-based models: a 3D–skeleton perception module to extract 3D perception from skeletons and a skeleton-image contrastive learning module to obtain the 2D–3D joint perception by fusing both multiview and skeleton perceptions. The experimental results reveal that MSFormer paired with large pre-trained multiview models achieves state-of-the-art performance, requiring only 100 training meshes. Furthermore, the segmentation accuracy is improved by 2.4\%–5.5\% with the increasing volume of training data.
\end{abstract}

\begin{IEEEkeywords}
mesh segmentation, dental objects, cross-modal methods, skeleton-multiview fusion
\end{IEEEkeywords}}
\maketitle

\IEEEdisplaynontitleabstractindextext

%% --------------------------------

\IEEEraisesectionheading{\section{Introduction} \label{sec:intro}}

\IEEEPARstart{T}{ooth} instance segmentation~\cite{task} serves as a fundamental task in dental computer–aided design (CAD) systems~\cite{inter_t3_curimportant, prahl1978need}. Precise instance segmentation is crucial for subsequent tasks, including biomechanical analysis~\cite{bm} and tooth arrangement~\cite{tar}. 
Given the high costs of manual tooth segmentation~\cite{darch}, there is a growing need to develop automated tooth segmentation methods. Although three-dimensional (3D) convolution-based~\cite{dseg,p3dinstance} and multiview-based methods~\cite{multiviewinstance,3dmv} have shown remarkable progress in instance segmentation; however, these methods encountered challenges in achieving optimal segmentation results, specifically for tooth segmentation.

3D convolution-based methods require extensive training data~\cite{qi2021review, huh2016makes}, as there is no available pre-trained model trained on large 3D instance segmentation datasets~\cite{deng2009imagenet}. Thus, acquiring copyrighted and labeled tooth data proves to be cost-prohibitive, thereby impeding the application of these methods to tooth segmentation~\cite{darch}. Lightweight models can be well trained with limited data; therefore, developing lightweight versions of these 3D convolution-based methods becomes a potential solution. However, these lightweight 3D models  often struggle to balance geometric details against full instance perception~\cite{sharp2022diffusionnet}, resulting in a coarse or poor segmentation.

\begin{figure}[htbp]
  \centering
  \includegraphics[width=\columnwidth,clip]{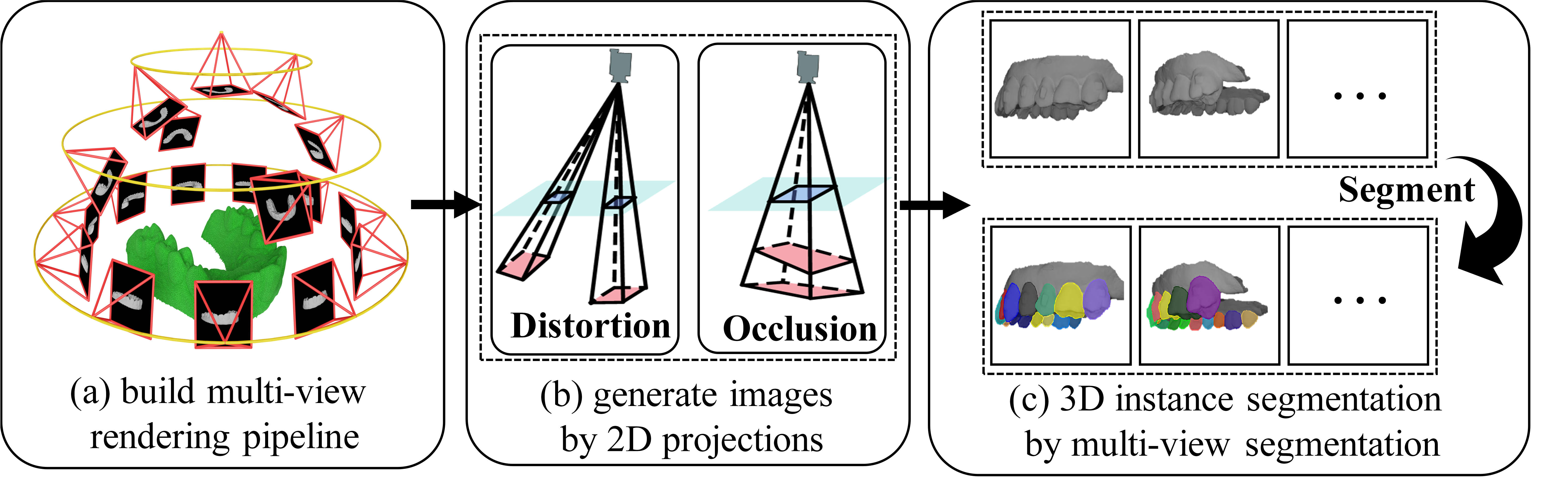}
  \caption{ Basic steps of multiview-based 3D instance segmentation. The 2D projection step is employed to convert 3D shapes into multiview images, and thereafter, the multiview-based methods implement tooth segmentation using these multiview images. }
\label{fig:2d_errors}
\end{figure}

\begin{figure}[htbp]  
  \centering
    \includegraphics[width=\columnwidth,clip]{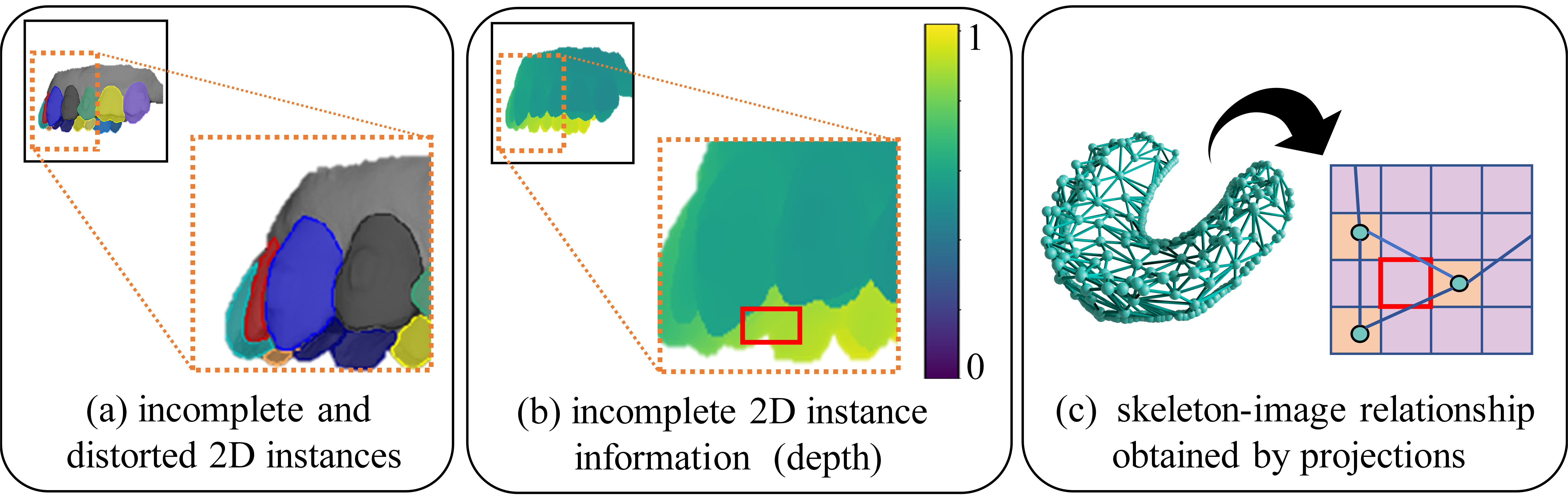}
    \caption{Problems raised by 2D projection. Owing to occlusion and distortion in 2D images (a,b), the biased and incomplete perception of instances is constructed in multiview-based methods. Moreover, the cross-modal relationship generated by 2D projection is incomplete. 2D image patches may require several adjacent skeleton nodes to overcome the biased perception; however, these nodes may not be projected into the image patch, such as the image patches \protect\redrect ~in (c). }
    \label{fig:othererrorfactor}
    \end{figure}

In contrast, multiview-based instance segmentation methods can leverage numerous pre-trained models trained on extensive two-dimensional (2D) instance segmentation datasets. These pre-trained models, rich in knowledge and prior support, enable effective fine-tuning, even with limited training data~\cite{liu2022petr,qi2021review}. However, these methods generally employ a 2D projection step, leading to complex issues, such as occlusion and distortion in multiview perception (Fig.~\ref{fig:2d_errors}). As shown in Fig.~\ref{fig:othererrorfactor}(a, b), these issues further result in incomplete and distorted perception of 3D shapes and suboptimal segmentation results.  Increasing the number of views by adding hundreds of images~\cite{hundredsimage} can alleviate this issue, but the increased massive computing resources make it not a promising solution.

The multiview-based method cannot achieve accurate segmentation results because of the biased and incomplete perception from 2D images.
 However, 3D convolution-based methods can provide a comprehensive perception of 3D shapes.
Therefore, integrating the 3D convolution-based methods into the multiview-based methods is a natural choice to address the biased and incomplete perception of multiview-based methods,
implying the development of a 2D–3D joint model. 
The main challenge in this joint model is achieving optimal segmentation results with limited training datasets,
 which means lightweight modules are more feasible than parameter-rich modules that require extensive training data. As lightweight modules struggle to balance the perception of shape outlines and geometric details, the input 3D data must be compacted and not involve too many geometric details. Thus, skeletons are a reasonable choice for 3D inputs. Following the above analysis, a new skeleton-multiview fusion method named MSFormer is proposed for tooth segmentation by integrating two lightweight modules into multiview-based models: a lightweight module named SkeletonNet aims to extract the comprehensive perception of 3D shapes, and a lightweight skeleton-multiview contrastive learning module aims to obtain 2D-3D joint perception by fusing multiview and skeleton perceptions.

 To ensure the effectiveness of these added modules, several novel designs are introduced. First, SkeletonNet introduces a new graph pooling operation to construct the hierarchical structures of skeletons, allowing for comprehensive perception of 3D shapes, even with limited convolution layers. Second, a lightweight skeleton-multiview contrastive learning loss is designed to facilitate learning a comprehensive cross-modal relationship between skeleton nodes and image patches. By learning this comprehensive relationship, the contrastive learning module can avoid many perception fusion failures resulting from incomplete cross-modal relationships, such as those highlighted in Fig.~\ref{fig:othererrorfactor}(c). Then, our contrastive learning module can obtain 2D-3D joint perception with a more robust performance.

 In summary, MSFormer combines 3D-skeleton perception with multiview perception to correct the biased and incomplete perception of multiview-based models by adding two lightweight modules. Our contributions can be summarized as follows:
\begin{itemize}
    \item We propose a skeleton-multiview fusion method, MSFormer, to optimize tooth segmentation with limited training data by adding two lightweight modules. Experimental results indicate that MSFormer delivers state-of-the-art (SOTA) performance with only 100 training meshes and improves the segmentation accuracy by more than 2\% as training data volume increases.
    \item We design SkeletonNet, which is a lightweight 3D perception module and capable of effective training with limited data to acquire 3D-skeleton perception.
    \item We present a lightweight skeleton-multiview contrastive learning module to integrate 3D skeleton and multiview perceptions, requiring limited training data and contributing to achieving optimal segmentation outcomes.
\end{itemize}

%% --------------------------------

\Section{Related work}
Tremendous efforts have been dedicated to tooth instance segmentation, which predominantly encompasses two principal categories: non-deep-learning-based methods and deep-learning-based methods.

\begin{figure*}[htbp]
  \centering
  \includegraphics[width=1.0\textwidth,trim={0 0 0 0},clip]{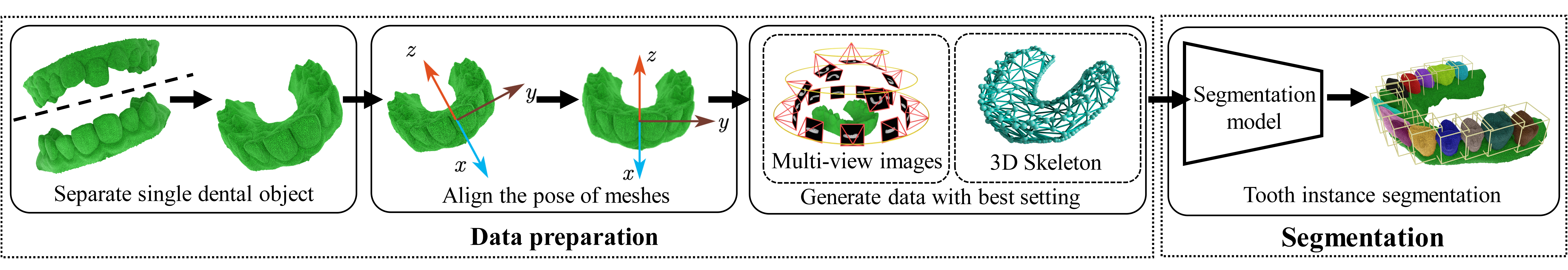}
  \caption{ Pipeline of MSFormer. First, we generate multiview images and skeletons of tooth meshes, and thereafter, employ a neural network to extract tooth instances from meshes.  }
  \label{fig:pipline}
\end{figure*}

\SubSection{Non-deep-learning-based methods}
Prior to the advent of deep learning, prevalent tooth segmentation methodologies were grounded in traditional machine learning techniques, bifurcating into expert knowledge-based methods and data-driven methods.

\textbf{Expert knowledge-based methods:} Initially constrained by technological and data limitations, the majority of research centered on interactive segmentation frameworks steered by expert knowledge~\cite{inter_t1,inter_t3_curimportant,inter4_courtour}. These methods rely on human–machine interactions to guide traditional machine learning–based segmentation methods for rapid labeling, thus expediting the manual segmentation process for most meshes~\cite{inter4_courtour}. However, the requisite to train experts for these interactive schemes often demands substantial effort, and the segmentation quality remains inconsistent. In contrast, MSFormer does not rely on these interactions in segmentation; thus, the significant efforts to master interactions can be avoided, and the inconsistent segmentation quality can be controlled.

\textbf{Data-driven methods:} With the maturation of data-driven machine learning algorithms, there is a growing interest in devising fully automated tooth instance segmentation methods to avoid unstable manual segmentation. These methods can be categorized into the 2D image–based and 3D mesh–based categories. On one hand, given the widespread application of multiview scanning in dental data acquisition, image-based segmentation methods, such as plane-view-based~\cite{image1,rangeimages1} and panoramic image–based methods~\cite{pano_images1}, serve as practical alternatives. However, due to constraints in data and computational resources, these 2D image–based methods cannot cover most tooth instances and can only offer rough segmentation results. On the contrary, 3D mesh–based methods aim to achieve tooth segmentation via mesh feature calculations, specifically targeting shape descriptors,  such as curvature field~\cite{planeview1}, contour-line~\cite{inter_t2}, and skeletons~\cite{skeletonimportant}, followed by face classification or clustering. Nevertheless, Chen et al.~\cite{benchmarkp} revealed the limitations of these methods, primarily their inability to adequately capture semantic information for mesh segmentation. MSFormer, built on deep-learning technologies, overcomes these limitations by capturing richer semantic information, thus yielding superior segmentation outcomes.

\SubSection{Deep-learning-based methods}
With the development of geometric deep learning~\cite{gdplearing}, a new array of instance segmentation methods based on deep learning has emerged. These methods can be classified into two distinct categories: general instance segmentation methods and tooth instance segmentation methods.

\textbf{General instance segmentation methods} principally include two types of methods: 3D convolution-based methods~\cite{xie2015projective} and multiview-based methods~\cite{multiviewinstance,3dmv}. As these general instance segmentation methods are not specifically tailored for tooth segmentation, they overlook many essential issues in tooth segmentation and suffer from suboptimal results. First, 3D convolution-based methods ignore the huge data preparation costs and require an extensive training dataset, making it prohibitively expensive. Second, many modules of existing multiview-based methods might not be suitable for tooth segmentation, such as the bird-eye-view-based modules for feature fusion~\cite{li2022bevformer,petr} and the voting modules between massive images for segmentation~\cite{xie2015projective}. Usually, Bird’s-eye-view-based modules include steps to flatten shapes, resulting in massive loss of shape information and making it more suitable for flat 3D scenes (e.g., road and indoor scenarios at the same horizontal plane) than tooth meshes. Additionally, voting modules between massive images improve segmentation results but usually lead to unacceptable computational resource requirements. Consequently, the forced application of these methods to tooth segmentation might not produce the desired segmentation results. Compared with these methods, MSFormer is specifically designed for tooth segmentation and can be well-trained with a limited data source. Unlike its general counterparts, MSFormer does not overlook these issues and achieves better segmentation results.

\begin{figure*}[ht!]
  \centering
  \includegraphics[width=1.0\textwidth,clip]{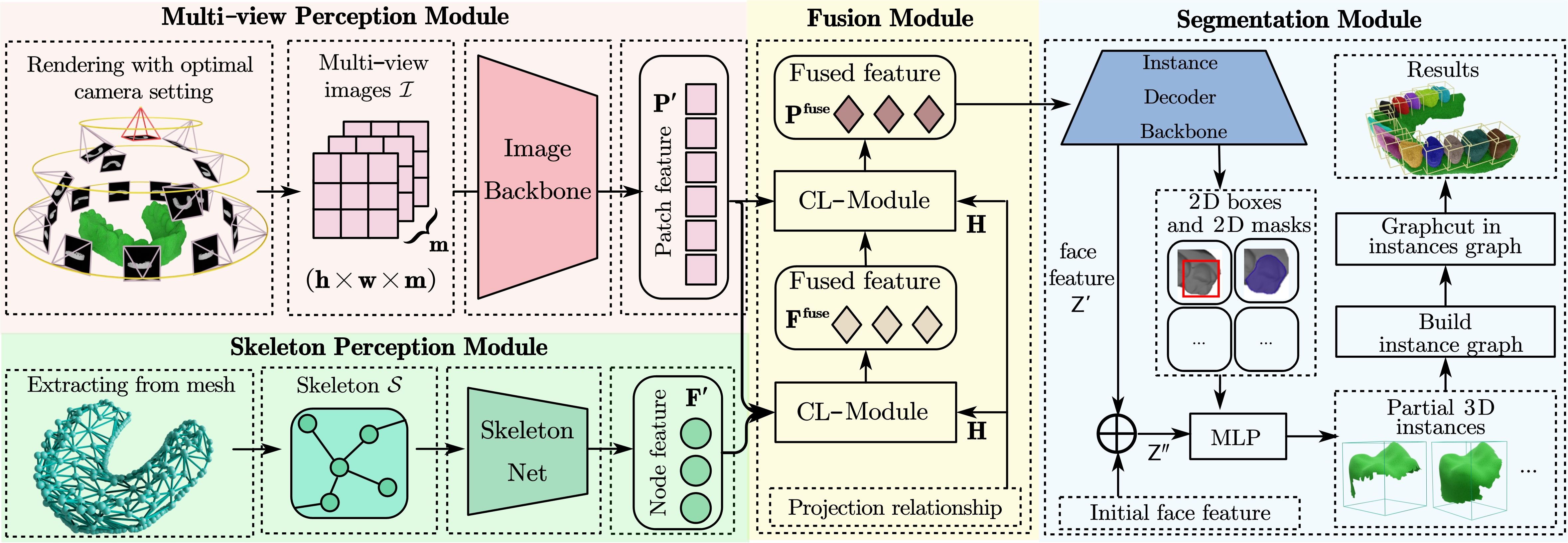}
  \caption{ Neural network architecture of MSFormer. It is composed of four modules: a multiview perception module for multiview feature extraction, a skeleton perception module for skeleton feature extraction, a fusion module for feature integration, and a segmentation module for tooth instance segmentation. Both the multiview perception module and segmentation module are pre-trained, whereas the skeleton perception module and fusion module are designed to be lightweight.}
  \label{fig:backbone}
\end{figure*}

\textbf{Tooth instance segmentation methods:} Recognizing the significance of prior knowledge specific to tooth meshes, researchers have developed a variety of instance segmentation methods~\cite{maskrnet,tsegnet}. For instance, Xu et al.~\cite{zheng1} emphasized the role of particular shape descriptors like minimum curvature~\cite{mincurvimportant} in tooth segmentation. These descriptors are converted into input images and subsequently processed using convolutional neural networks to achieve the desired segmentation. Recently,  a significant body of research has been devoted to developing ad-hoc convolution methods on point cloud for enhanced segmentation accuracy, such as Monte Carlo convolution methods~\cite{maskrnet} and two-head-predictor~\cite{teethgnn}. The main criticism of these methods is that these ad-hoc methods usually contain many non-pretrained parameters, necessitating large volumes of training data. To ameliorate this issue, weakly supervised methods~\cite{darch} have been proposed to mitigate the labeling costs. However, these methods continue to require substantial training data and incur significant data collection expenses. In comparison, MSFormer reduces both the labeling and data-collecting costs by integrating two lightweight modules to rectify the biased perception of multiview-based methods. MSFormer can be well-trained using a limited dataset and obtain optimal segmentation results. Furthermore, MSFormer uses tooth shape priors to determine the optimal camera setting and render images while not involving inappropriate priors about tooth sizes and positions. Consequently, MSFormer successfully circumvents issues that commonly plague other methods, such as insufficient cropping~\cite{darch,tsegnet}.

%% --------------------------------

\Section{MSFormer}
\label{sec:method}
As illustrated in Fig.~\ref{fig:pipline}, MSFormer commences by generating multiview images and skeletons of tooth meshes as inputs (Sec.~\ref{sec:datap}), subsequently executing instance segmentation via a neural network (Sec.~\ref{sec:msformer}).

\SubSection{Data preparation}
\label{sec:datap}
A comprehensive dental object comprises meshes for both the upper and lower jaws. Undertaking segmentation based on multiview images of the entire dental object exacerbates occlusion. Thus, segmentations are performed separately on the upper and lower tooth meshes. To neutralize the impact of varying postures, we employ principal component analysis to align the postures of tooth meshes, consistent with prior research~\cite{zheng1}. Once the tooth meshes are aligned in terms of posture, the specific procedures for rendering multiview images and extracting skeletons are explained below.

\SubSubSection{Multiview image rendering}
The rendering process for multiview images encompasses two primary steps. Initially, a multiview camera setting is established to facilitate the rendering. 
As explained in the appendix, we identify the optimal camera setting $\mathcal{V} = \{V_{1}, ..., V_{i},..., V_{m}\}$, which uses the fewest number $m$ of views to cover the most area of tooth instances (over 99.9\% vertices of each tooth instance averagely). Subsequently, mesh $M$ is rendered from each camera $V_{i} \in \mathcal{V}$, yielding a series of multiview images  $\mathcal{I} = \{I_{1}, ..., I_{i},..., I_{m}\}$, where $I_{i} $ represents a $512 \times 512 $ image, and $m=16$ in this paper.

\SubSubSection{Skeleton extraction of meshes}
Several algorithms exist for skeleton extraction from meshes, including mesh contraction~\cite{skeletonextract}, mesh segmentation~\cite{skeletonextract1}, and medial axis transformation~\cite{meidal}. However, these algorithms often depend on intricate calculations and closed manifold surfaces, rendering them highly unstable for tooth meshes characterized by many topological errors. Consequently, this paper adopts the spectral clustering algorithm~\cite{spectral} for generating tooth skeletons, which do not rely on complex and fragile computation on closed manifold surfaces.

Given a single tooth mesh $M$, spectral clustering performed on its faces yields a tooth skeleton $\mathcal{S}= (\mathcal{N}, \mathcal{E})$, where $\mathcal{N}$ denotes a set of skeleton nodes with size  $n$, and $\mathcal{E}$  denotes a set of skeleton edges. Each skeleton node $N_{i}$ corresponds to a cluster of faces, and an edge is introduced between two skeleton nodes in $\mathcal{E}$ if their respective face clusters are adjacent. To initialize face features, we compute widely used shape descriptors, including coordinates, normal vectors, shape diameter, fast point feature histograms, and four types of curvature (mean, maximum, minimum, and gaussian).  The initial feature $F$ of skeleton nodes is determined by aggregating the features of faces within the corresponding clusters.

In this manner, utilizing the optimal camera setting, we establish the multiview rendering pipeline and procure a set of multiview images $\mathcal{I}$. Additionally, by applying the spectral clustering algorithm to the meshes, we ascertain the skeleton $\mathcal{S}$ of each mesh.

\begin{figure*}[htbp]
    \centering
    \includegraphics[width=1.0\textwidth,trim={2 2 2 2},clip]{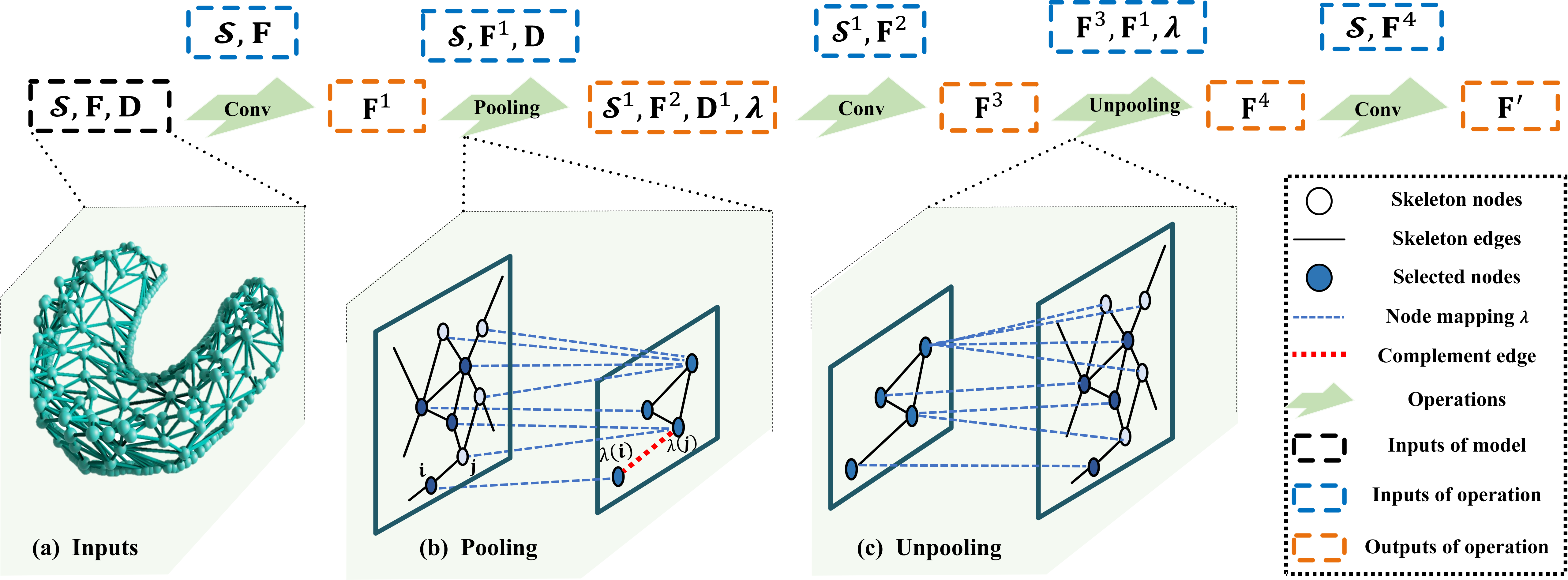}
    \caption{
    Architecture of SkeletonNet. The input of each operation is the data in the upper box, and the output is the data in the right box. (a) represents the input skeleton. (b) represents the pooling operation. The downsampled skeleton is constructed by removing unimportant nodes, and  the node mapping $\lambda$ is used to complement edges and ensure the connectivity of the downsampled skeleton, such as the added edge between  $\lambda(i) \to \lambda(j)$. (c) is the unpooling operation integrating the node features of downsampled skeletons with the original features via the node mapping $\lambda$.
}
    \label{fig:skeletonNet}
\end{figure*}

\SubSection{Segmentation}
\label{sec:msformer}
Upon obtaining the multiview images $\mathcal{I}$ and the skeleton  $\mathcal{S}$  of mesh $M$, MSFormer accomplishes tooth segmentation by synthesizing perceptions from both the skeleton and the multiview images, as portrayed in Fig.~\ref{fig:backbone}. The neural network architecture of MSFormer is compartmentalized into four distinct modules: a \textbf{multiview perception module} designed to extract features from multiview images; a \textbf{skeleton perception module} focused on extracting node features from skeletons; a \textbf{fusion module}  responsible for the amalgamation of multiview image features with skeleton node features; and a \textbf{segmentation module} tasked with carrying out instance segmentation. The details of these modules are elaborated upon in the following sections.

\SubSubSection{Multiview perception module}
The multiview perception module utilizes an image encoder to extract features from multiview images. In this paper, we adopt existing pre-trained image feature extractors, such as Swin-Transformer~\cite{swin} and ViT~\cite{eva}, as our encoder. Given a set of multiview images $\mathcal{I}$, the entire multiview perception module can be expressed as follows:
\begin{equation}
    \label{equ:multiviewmodule}
    P = \bigoplus_{I_{i} \in \mathcal{I}} Encoder(I_{i}),
\end{equation}
where $Encoder$ represents a pre-trained image feature extractor, $\bigoplus$ represents the feature concatenation operation,  $P \in \mathbb{R}^{m \times h \times w \times c}$ represents image patch features of $m$ multiview images, $h \times w$ denotes the number of patches for a single image, and $c$ indicates the dimension of features for each image patch.   Reshaping $P$ as $P' \in \mathbb{R}^{p \times c}$, we obtain a list of image patch features, where $p = m \times h \times w $ is the total number of patches.

\SubSubSection{Skeleton perception module}
Owing to the relatively compact dimensions of skeletons, a limited number of convolution layers suffice to encapsulate the full perceptions of 3D shapes. To further economize on the convolutional parameters, we employ graph pooling algorithms~\cite{psurvey} for the extraction of skeleton node features. Essentially, these graph pooling methods establish hierarchical structures of skeletons, thereby requiring fewer convolutions on these hierarchies to attain a comprehensive perception of 3D shapes.

Two prevalent categories of graph pooling algorithms are node drop pooling and node clustering pooling~\cite{psurvey}. Of these, node drop pooling algorithms are distinguished by their stability, computational efficiency, and robust generalization capabilities~\cite{graphUnet}, rendering them particularly suitable for tasks with limited data sets. Nevertheless, a critical drawback associated with node drop pooling algorithms is the potential disruption of skeletal connectivity during the construction of hierarchical structures. As shown in Fig.~\ref{fig:skeletonNet} (b), a pooling operation relying exclusively on node dropping may produce isolated components, thereby impeding comprehensive shape perception. To mitigate this concern, the present study introduces a novel lightweight module, dubbed SkeletonNet, which is designed to extract the comprehensive 3D shape perception while preserving skeletal connectivity during the node-dropping process. As illustrated in Fig.~\ref{fig:skeletonNet}, SkeletonNet comprises three fundamental operations: \textbf{Conv}, \textbf{Pooling}, and \textbf{Unpooling}. The inputs of each operation are one skeleton $\mathcal{S}^{in} = (\mathcal{N}^{in}, \mathcal{E}^{in})$, the skeleton node features $F^{in}$, and the pre-computed geodesic distance matrix $D^{in}$, where 
$\mathcal{N}^{in}$ is the set of skeleton nodes with size $n$, and  $\mathcal{E}^{in}$ is the set of edges. The details can be described as follows.

\textbf{Conv}. The $Conv$ operation is geared toward achieving shape perception through skeletal convolution.  Given one skeleton $\mathcal{S}^{in}$ and the node feature  $F^{in} \in \mathbb{R}^{n \times l}$, the convolution operation $Conv^{edge}$ of any edge $(i \rightarrow j) \in \mathcal{E}^{in}$ can be defined as:
\begin{equation}
    \label{equ:convedge}
    Conv^{edge}(i \rightarrow j) = GELU(\gamma(\gamma(F^{in}_{i}) - \gamma(F^{in}_{j}))) \cdot \gamma(F^{in}_{j}),
\end{equation}
where $n$ denotes  the node size of the skeleton, $l$ indicates  the dimension of node features,  $F^{in}_{i} \in F^{in}$ is the node feature of $i$-th node, $F^{in}_{j} \in F^{in}$ represents  the node feature of $j$-th node, each $\gamma$ is a learnable linear projection from feature dimension $l$ to $l$, and $GELU$ represents  an activate function~\cite{gelu}. Therefore, the convolution operation $Conv^{node}$ on $j$-th node can be defined as:
\begin{equation}
    \label{equ:convnode}
    Conv^{node}(j) = \sum_{j' \in neighbor(j)}  Conv^{edge}(j' \rightarrow j),
\end{equation}
where $neighbor(j)$ represents the adjacent nodes of $j$-th node. Applying the operation of skeleton convolution to all skeleton nodes, the shape perception can be embedded into the updated node feature $F^{out} \in \mathbb{R}^{n \times l}$:
\begin{equation}
    \label{equ:conv}
    F^{out} = \bigoplus_{j \in \mathcal{N}^{in}} Conv^{node}(j),
\end{equation}
where $\bigoplus$ means a feature concatenation operation. 
Accordingly, given one skeleton $\mathcal{S}^{in}$ and node feature $F^{in}$, the $Conv$ operation on skeletons can be summarized as follows:
\begin{equation}
    \label{equ:sconv}
     F^{out} = Conv(\mathcal{S}^{in}, F^{in})
\end{equation}

\textbf{Pooling.} 
Utilizing single skeleton $\mathcal{S}^{in}$, the node features $F^{in} \in \mathbb{R}^{n \times l}$, and the geodesic distance matrix $D^{in} \in  \mathbb{R}^{n \times n}$ as inputs, the pooling operation aims to derive a downsampled skeleton. This operation can be subdivided into four steps.
\emph{First}, the attention scores of the self-attention mechanism~\cite{vaswani2017attention} served as the importance score  $C \in \mathbb{R}^{n}$ of the nodes:
\begin{equation}
\label{equ:score}
C = self{\text -}attention(F^{in}).
\end{equation}
\emph{Second},  a $Topk$ operation is employed to select the $(r*n)$ nodes  with the highest importance scores: 
\begin{equation}
    \label{equ:topk}
     \mathcal{N}^{out} = Topk(C, r*n),
\end{equation}
where $r$ denotes  the pooling rate, i.e, the percentage of selected nodes, $\mathcal{N}^{out}$ indicates  a set of selected nodes with the highest importance scores. 
\emph{Third}, we construct a mapping $\lambda$ between skeleton nodes $\mathcal{N}^{in}$ and selected nodes $\mathcal{N}^{out}$ based on the geodesic distance  matrix $D^{in}$. 
Specifically, the node mapping $\lambda$  associates any selected node to itself and maps any unselected nodes to the nearest selected node in geodesic distance, as shown in Fig~\ref{fig:skeletonNet}(b). 
\emph{Finally}, converting each edge $ (i \to j) \in \mathcal{E}^{in}$ to $(\lambda(i) \to \lambda(j)) \in \mathcal{E}^{out}$, we obtain the downsampled skeleton $\mathcal{S}^{out} = (\mathcal{N}^{out}, \mathcal{E}^{out})$.  
Aggregating each node feature $F^{in}_{i}$ to the selected node' feature $F^{out}_{\lambda(i)}$ following the averages, we obtain the node features $F^{out}$ of the selected nodes. Aggregating $i$-th row of  $D^{in}$ and the $i$-th column to $\lambda(i)$-th row  and $\lambda(i)$-th column following the minimum, respectively, we obtain the new geodesic distance matrix $D^{out} \in \mathbb{R}^{(r*n) \times (r*n)}$ between selected nodes.  Thereafter,  the pooling operation can be summarized as follows:
\begin{equation}
    \label{equ:pool}
    \mathcal{S}^{out}, F^{out}, D^{out}, \lambda = Pooling(\mathcal{S}^{in}, F^{in}, D^{in}), 
\end{equation}
where the downsampled skeleton is $\mathcal{S}^{out} = (\mathcal{N}^{out}, \mathcal{E}^{out})$, $F^{out}$ and $D^{out}$ denote the updated node feature and geodesic distance matrices, respectively.

\textbf{Unpooling.} Given the skeleton feature $F^{ap}$ after pooling and the skeleton feature $F^{in}$ before pooling,  the unpooling operation aims to integrate $F^{in}$ and $F^{ap}$ based on node mapping  $\lambda$, such as $F^{3}$ as $F^{ap}$ and $F^{1}$ as $F^{in}$ in Fig.~\ref{fig:skeletonNet}. Specifically, for each node feature $F^{in}_{i} \in F^{in}$, we execute the following step:
\begin{equation}
    \label{equ:unpooling}
    F^{out}_{i} = F^{in}_{i} + F^{ap}_{\lambda(i)}, 
\end{equation}
where $F^{out}_{i}$ denotes the feature of the $i$-node after unpooling. Applying Eq.~\ref{equ:unpooling} to all node features $F^{in}$, the unpooling operation updates the node features:
\begin{equation}
    \label{equ:unpool}
    F^{out} = Unpooling(F^{ap}, F^{in}, \lambda), 
\end{equation}
where the fused feature $F^{out} \in \mathbb{R}^{n \times l}$ has the same shape as $F^{in}$.

\textbf{Architecture of SkeletonNet.} 
Given a skeleton $\mathcal{S}$, the initial node features $F \in \mathbb{R}^{n \times l}$, and the geodesic distance matrix $D \in  \mathbb{R}^{n \times n}$ between skeleton nodes, we employ the aforementioned three operations to formulate a streamlined model known as SkeletonNet. This model is designed to extract a comprehensive perception of 3D shapes, denoted by the node feature $F^{'}$ in Fig.~\ref{fig:skeletonNet}. Notably, SkeletonNet introduces no additional trainable parameters beyond those inherent in the existing node drop pooling method~\cite{graphUnet}. Moreover, SkeletonNet employs node mapping $\lambda$ to supplement edges and avoid connectivity changes  during the pooling phase, as exemplified by  the edge $(\lambda(i) \to \lambda(j))$ in  Fig.~\ref{fig:skeletonNet}(b). Therefore, SkeletonNet maintains the original skeleton's connectivity and precludes the formation of isolated components. This ensures comprehensive shape perception as well as enhances segmentation results when compared to general node drop pooling algorithms.  In this paper, we set the skeleton node number $n$ as 400 and the pooling rate $r$ as 0.4, according to the hyper-parameter analysis in Sec~\ref{sec:hyper}.

\SubSubSection{Fusion module}
Given the multiview features $P'  \in \mathbb{R}^{p \times c}$ and skeleton features $F' \in \mathbb{R}^{n \times l}$, the fusion module aims to obtain the 2D-3D joint perception by integrating these two cross-modal features. The fundamental challenge lies in capturing a comprehensive cross-modal relationship between $P'$ and $F'$ and subsequently executing feature fusion predicated on this relationship. Using the incomplete projection relationship $H \in \mathbb{R}^{p \times n}$ between image patches and skeleton nodes directly is not a good idea, as the feature fusion failure raised by an incomplete relationship may lead to the suboptimal segmentation results based on  single-modal perceptions. 

To avoid the incomplete cross-modal relationship, we develop a lightweight contrastive learning module, denoted as $CL\text{-}Module$, to discern a comprehensive cross-modal relationship and facilitate subsequent feature fusion. In alignment with previous research~\cite{clip}, 
\begin{figure}[ht!]
    \centering
    \includegraphics[width=1.0\columnwidth,trim={4 4 4 4},clip]{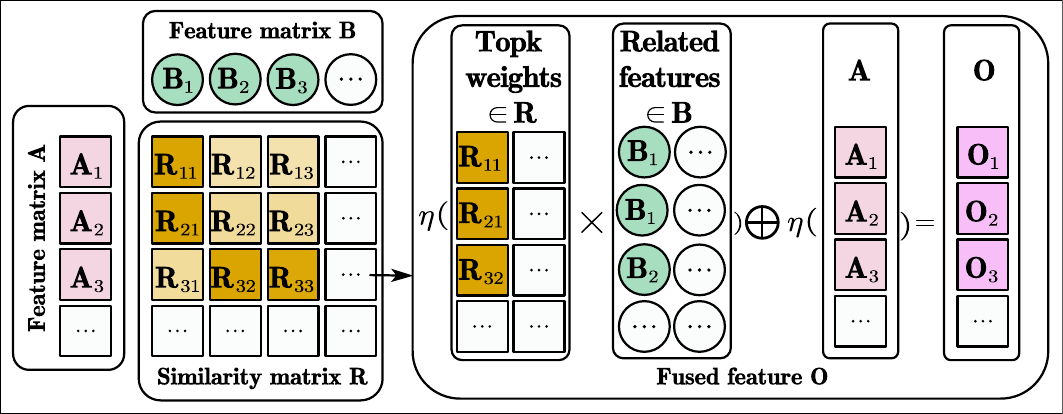}
    \caption{Feature fusion progress of contrastive learning module $CL{\text -}Module$ to integrate $B$ to $A$.}
    \label{fig:fusedfeature}
  \end{figure}
\begin{figure*}[ht!]
    \centering
    \includegraphics[width=1.0\textwidth,trim={4 4 4 4},clip]{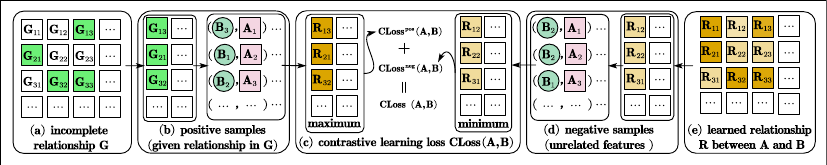}
    \caption{Definition of the contrastive learning loss $CLoss(A, B)$. Darker squares represent higher weights in the given incomplete relationship $G$ and the learned relationship $R$. 
    Essentially, $CLoss$ serves two purposes: minimizing the similarity between cross-modal features in negative samples (i.e., unrelated features), and maximizing the similarity between  cross-modal features in positive samples (i.e., given relationship $G$).}
    \label{fig:closs}
  \end{figure*}
the $CL\text{-}Module$ accomplishes cross-modal feature fusion through several trainable linear transformations, i.e., a weighted average operation. The main contribution of $CL\text{-}Module$ is a new contrastive learning loss,  designed to guide the learning of comprehensive cross-modal relationships using the pre-existing incomplete  projection relationship $H$ rather than learning from voluminous training data directly.

Given any two cross-modal features $A  \in \mathbb{R}^{a \times a'}$ and $B  \in \mathbb{R}^{b \times b'}$, and  an incomplete relationship $G \in \mathbb{R}^{a \times b}$ between $A$ and $B$,  the \textbf{feature fusion progress} and \textbf{contrastive learning loss} of $CL\text{-}Module$ can be defined as follows.

\textbf{The feature fusion progress} uses a weighted average operation to integrate $B$ to $A$, as summarized in Fig.~\ref{fig:fusedfeature}. This progress comprises three steps. 
Firstly, similar to previous contrastive modules~\cite{clip}, we learn the cross-modal relationship $R \in \mathbb{R}^{a \times b}$ between $A$ and $B$ as follows:
\begin{equation}
    \label{equ:allproj}
    R = matmul(\sigma(A), \tau(B)),
\end{equation}
where $matmul$ represents the matrix multiplication, $\sigma$ and $\tau$ represent trainable linear transformations to transform both $A$ and $B$ into the same dimension $d$, and  $R_{ij} \in R$ represents the learned relationship weight between  $A_i \in A$ and  $B_{j} \in B$, i.e., the learned similarity. 
Secondly, for feature $A_{i} \in A$, we  apply a weighted average operation to $k$ features in $B$ having the highest relationship weights to $A_{i}$:
\begin{equation}
    \label{equ:simi}
    A^{'}_{i} = \sum_{j \in \mathcal{T}} B_{j} \cdot  R_{ij}, 
\end{equation}
where  $\mathcal{T} = Topk(R_i, k)$ is  the indices of $k$ highest  relationship weights between $A_i$  and all features in $B$, and $A^{'}_{i}$ denotes the weighted average of $k$ features in $B$. 
Finally, by concatenating $A^{'}_{i}$ and $A_{i}$, we integrate $B$ to $A_{i}$:
\begin{equation}
    \label{equ:out}
    O_{i} = \eta( A^{'}_{i}) \oplus \eta(A_{i}), 
\end{equation}
where $O_{i} \in \mathbb{R}^{1 \times d}$ is the fused feature, each $\eta$ is a trainable linear transformation employed to reduce dimension $d$ to $d/2$, and $\oplus $ represents a feature concatenation operation. Applying the above steps to all features in $A$, $CL\text{-}Module$  obtains the fused feature  $O \in \mathbb{R}^{a \times d}$ by integrating $B$ to $A$.

\textbf{The contrastive learning loss} capitalizes on the incomplete relationship $G$ to both guide the learning of comprehensive cross-modal relationship $R$ and mitigates the need for extensive training data. This is illustrated in Fig.~\ref{fig:closs}. The loss function itself can be bifurcated into two components: loss derived from positive samples and loss derived from negative samples.

\emph{The loss derived from positive samples} ensures that the existing cross-modal relationship  $G$, despite being incomplete, is  included in the learned relationship $R$ without necessitating large volumes of training data. Specifically, the elements of $G$ serve as positive samples, and the corresponding loss aims to maximize the weights of these positive samples in the learned relationship. For example, for any given relationship $G_{it} > 0$ between feature $A_i$ and feature $B_t$, we aim for a sufficiently large $R_{it}$ to be of a magnitude such that it falls within the top $k$ largest relationship weights of $R_{i}$. This ensures that the given relationship $G_{it}$ will be considered in feature fusion. Thus, for  any $A_i \in A$, the loss $CLoss^{pos}(A_i, B)$ derived from positive samples can be written as follows:
\begin{equation}
    \label{equ:pcloss}
\begin{split}
    CLoss^{pos}(A_i, B)  = \sum_{\{t | G_{it} > 0\}} max(min(\{R_{ij} | j \in \mathcal{T}\})  \\
    - R_{it}, 0 ),
\end{split}
\end{equation}
where $\{t | G_{it} > 0\}$ represents the given incomplete relationship in $G$, and $min(\{R_{ij} | j \in \mathcal{T}\})$ denotes the minimum relationship weight  in $k$ highest relationship  weights. According to Eq.~\ref{equ:pcloss}, $CLoss^{pos}(A_i, B)$ will be positive if any given cross-modal relationship is not considered in feature fusion. Thus, optimizing $CLoss^{pos}(A_i, B)$ to zero will ensure that given relationship $G$ can be included into the learned relationship $R$ without massive training data.

\emph{The loss derived from negative samples } ensures that the learned relationship will  not be excessively redundant. If the learned relationship is redundant, unrelated features might have high relationship weights in $R$, which may interfere with the feature fusion. Therefore,  the relationship without the top $k$ largest relationship weights will be regarded as negative samples. The loss derived from negative samples aims to minimize the relationship weights of these negative samples (i.e., unrelated features) as much as possible. For each feature $A_i \in A$, the loss  $CLoss^{neg}(A_i, B)$  derived from negative samples can be defined as:
\begin{equation}
    \label{equ:ncloss}
    CLoss^{neg}(A_i, B)  = 1 - \sum_{j \in \mathcal{T}} R_{ij}, 
\end{equation}
where  $\sum_{j \in \mathcal{T}} R_{ij}$ represents the sum of $k$ highest relationship weights between $A_i$ and all features in $B$. Optimizing $CLoss^{neg}(A_i, B)$ to zeros will  minimize the learned relationship weighted between unrelated features, which makes the learned relationship non-redundant excessively.

We use the sum of $CLoss^{pos}$ and $CLoss^{neg}$  as the contrastive learning loss $CLoss$, so the learned cross-modal relationship $R$ will include the incomplete relationship $G$ but not be excessively redundant.  Applying the $CLoss$ to all features in $A$, we obtain the contrastive loss $Loss_{cons} = CLoss(A, B)$. 
In this way, the contrastive learning module $CL{\text -}Module$, including this feature fusion progress and contrastive learning loss, can be summarized as follows:
\begin{equation}
    \label{equ:cl}
    O, Loss_{cons} = CL{\text -}Module(A, B, G, k),
\end{equation}
where $O$ is the fused feature between cross-modal $A$ and $B$,  $Loss_{cons}$ represents the contrastive loss.  
\begin{figure*}[ht!]
    \centering
    \includegraphics[width=0.95\textwidth,clip]{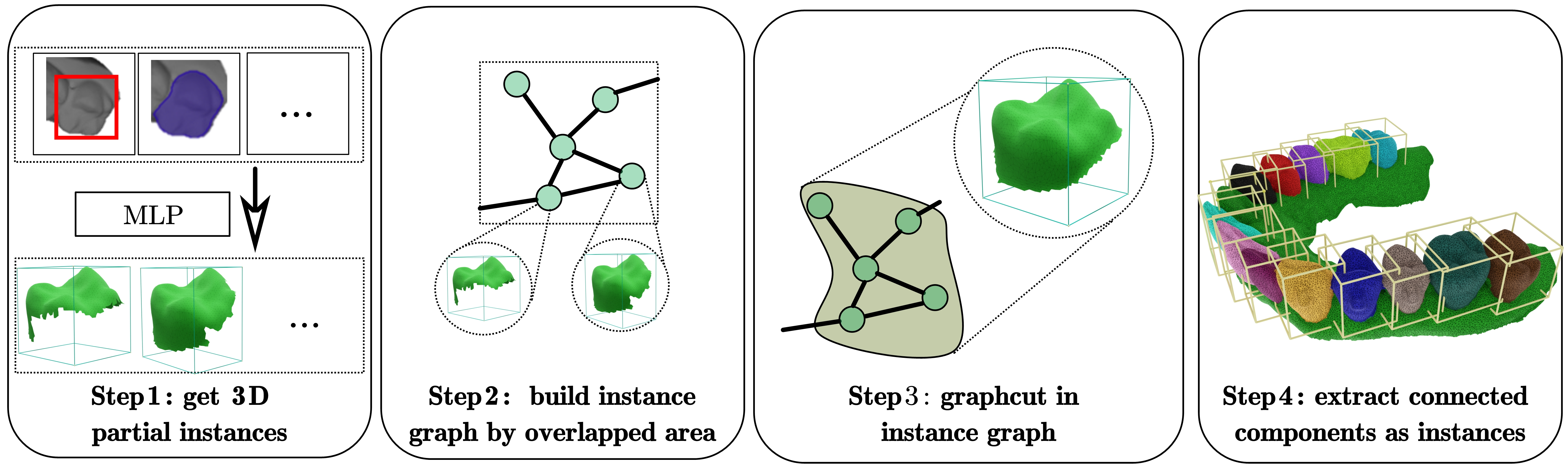}
    \caption{Steps of graphcut to merge partial instances to complete tooth instances.}
    \label{fig:graphcut}
\end{figure*}

As shown in Fig.~\ref{fig:backbone}, the fusion module comprises two $CL\text{-}Module$ . Using the projection relationship $H$ as guidance, the first  $CL\text{-}Module$  (Eq.~\ref{equ:fuse1}) obtains the fused features $F^{fuse}$ by integrating multiview features $P'$ to skeleton features $F'$, and the second one (Eq.~\ref{equ:fuse2}) obtains the enhanced multiview features $P^{fuse}$ by integrating the fused features $F^{fuse}$ to  multiview feature $P'$  as follows:
\begin{align}
    F^{fuse}, Loss^{1}_{cons}  = CL{\text -}Module(F', P', H, k * m),  \label{equ:fuse1}\\
    P^{fuse}, Loss^{2}_{cons}  = CL{\text -}Module(P', F^{fuse}, H, k * m), \label{equ:fuse2}
\end{align}
Accordingly, this fusion module obtains the enhanced multiview features $P^{fuse}$ by fusing multiview perception and skeleton perception. 

\SubSubSection{Segmentation module}
By inputting the enhanced multiview features $P^{fuse}$ into the segmentation module, we can obtain the instance segmentation results of all views:
\begin{equation}
\label{equ:seg}
\mathcal{Z} = Decoder(P^{fuse}),
\end{equation}
where $Decoder$ represents the pre-trained segmentation backbone, and  $\mathcal{Z} = \{Z_{1}, ..., Z_{i}, ..., Z_{q}\}$ represents the pixel feature of $q$ instances. 
Furthermore, pixel features are back-projected onto faces, and face features  $\{Z^{'}_{1}, ..., Z^{'}_{i}, ..., Z^{'}_{q}\}$ of $q$ instances are generated by concatenating these pixel features with initial face features—specifically, i.e., pre-computed shape descriptors. Upon standardizing the face features of each instance, these normalized features are input into a multilayer perceptron (MLP) to predict the labels for individual faces. Eliminating faces labeled as gum, we obtain the instances with only tooth parts. The overall training loss for MSFormer can be divided into two parts: the loss related to mispredicted facial areas, $Loss_{area}$, and the contrastive learning loss, $ Loss^{1}_{cons} + Loss^{2}_{cons}$.

Given the potentially incomplete nature of input instances derived from images, the outputs of MLP are often partial tooth instances without gum parts, as illustrated in Fig.~\ref{fig:graphcut}. Therefore, we must merge these partial tooth instances into complete tooth instances. A key observation here is the significant overlap among partial instances when they belong to the same tooth, whereas such instances remain isolated when they belong to different teeth. Utilizing this observation, we construct an instance graph and apply a graph cut approach~\cite{gcut} to merge these partial instances, as illustrated in Fig.~\ref{fig:graphcut}. Specifically, each partial instance is treated as an individual node; edges are constructed between overlapping instances, with the magnitude of overlap serving as the edge weight. As each tooth only appears once in an image, the instances within a single image correspond to different teeth. Therefore, merging partial instances within the same image into a connected component is deemed an error in the graph cut methodology. Given that the number of teeth in a tooth mesh is not expected to exceed 20, an initial cut threshold is set to segment the mesh into 21 connected components. This threshold is subsequently adjusted downward to yield fewer connected components. The minimum cut threshold with the least number of errors is selected for the final segmentation results.

%% --------------------------------
\Section{Experiments}
\label{sec:exp}

\begin{table*}[htbp]
  \centering
  \caption{Statistical information of our dataset, including number of vertices, ratios of adjacent teeth, and percentages of non-manifold vertices. The ratio of adjacent teeth is determined by dividing the face count of the larger tooth by that of the smaller tooth.}
    \begin{tabular}{lcccccccccc}
    \toprule
    \multicolumn{1}{c}{\multirow{2}[4]{*}{Dataset}} & \multicolumn{3}{c}{Number of vertices} & \multicolumn{3}{c}{Ratios of adjacent teeth} & \multicolumn{3}{c}{Non-manifold vertices (\%)} & \multirow{2}[4]{*}{\parbox{1.5cm}{Number of meshes}} \\
  \cmidrule(r){2-4}\cmidrule(r){5-7}\cmidrule(r){8-10}
  & Min   & Mean  & Max   & Min   & Mean  & Max   & Min   & Mean  & Max   &  \\
    \midrule
    Tooth(upper) & 20,692 & 125,180 & 385,180 & 1.01  & 1.98  & 22.96 & 0     & 0.7   & 4.7   & 4,000 \\
    Tooth(lower) & 17,594 & 139,444 & 409,778 & 1.00  & 1.86  & 19.92 & 0     & 1.1   & 3.1   & 4,000 \\
    \bottomrule
    \bottomrule
    \end{tabular}%
    \label{tab:dataset}%
\end{table*}%

\SubSection{Dataset preparation}

\textbf{Source Dataset.}  We build a source dataset by gathering 8,000 tooth meshes from 3,769 patients. This dataset comprises 4,000 upper teeth, 4,000 lower teeth, and 123,312 tooth instances.  We invited 30 dentists for precise manual annotation to curate the original 3D data and create segmentation labels. As shown in Tab.~\ref{tab:dataset}, our dataset comprises various real-world meshes, including those with dental deformities and topology errors. Segmenting these real-world meshes poses a significant challenge, as many existing methods rely on assumptions that may not be fit for these real-world meshes. For example, some tooth segmentation methods are based on three-step progress:  detecting each tooth instance, cropping a fixed-sized 3D patch per instance, and segmenting the foreground tooth within each 3D patch. However, in cases where a 3D patch contains both a large foreground tooth and adjacent small teeth, these methods may inadvertently overlook the small teeth, leading to reduced segmentation accuracy. 
The ratios of adjacent teeth in Tab.~\ref{tab:dataset} suggest that 3D patches containing one large tooth and neighboring small teeth are a regular occurrence in our dataset.   Consequently, the real-world complexity of these meshes poses substantial challenges for methodological robustness.

\textbf{Multiview images and skeletons Generation.} Initially, we generate 16 multiview images for each tooth mesh, each with dimensions of $512 \times 512$ pixels, in accordance with the optimal camera settings $\mathcal{V}$ in Sec.~\ref{sec:datap}. During the training phase, data augmentation is performed by applying random rotations within an angular range of 0 to $\pi/12$ radians to ensure that MSFormer remains robust to minor rotational variations. Subsequently, spectral clustering is applied to the meshes to generate skeletons comprising 400 nodes. The identical camera setting $\mathcal{V}$ is utilized to establish the projection relationship between the skeletal nodes and image patches. This process yields multiview images, skeletons, and their corresponding projection relationships.

\textbf{Training dataset and testing dataset.} To evaluate segmentation performance across diverse dataset sizes, $i$ meshes are utilized for training while the remaining $(8,000 - i)$ meshes serve as the test set. The variable $i$ assumes values of 100, 500, 2,000, and 4,000, respectively. Specifically, both upper and lower tooth meshes are divided into $ 8,000/i$ equal parts, respectively. The training dataset comprises one part from the upper teeth and one from the lower teeth; the remaining parts constitute the testing dataset. By repeating evaluations $8,000/i$ times, the mean accuracy gleaned from these tests serves as the final test accuracy. All experimental evaluations are conducted on a single computational system equipped with four RTX 3090 GPUs and an Intel(R) Xeon Gold 6133 CPU.

\SubSection{Experimental setup}

\begin{figure*}[ht!]
  \centering
  \includegraphics[width=1.0\textwidth,trim={8 8 8 8},clip]{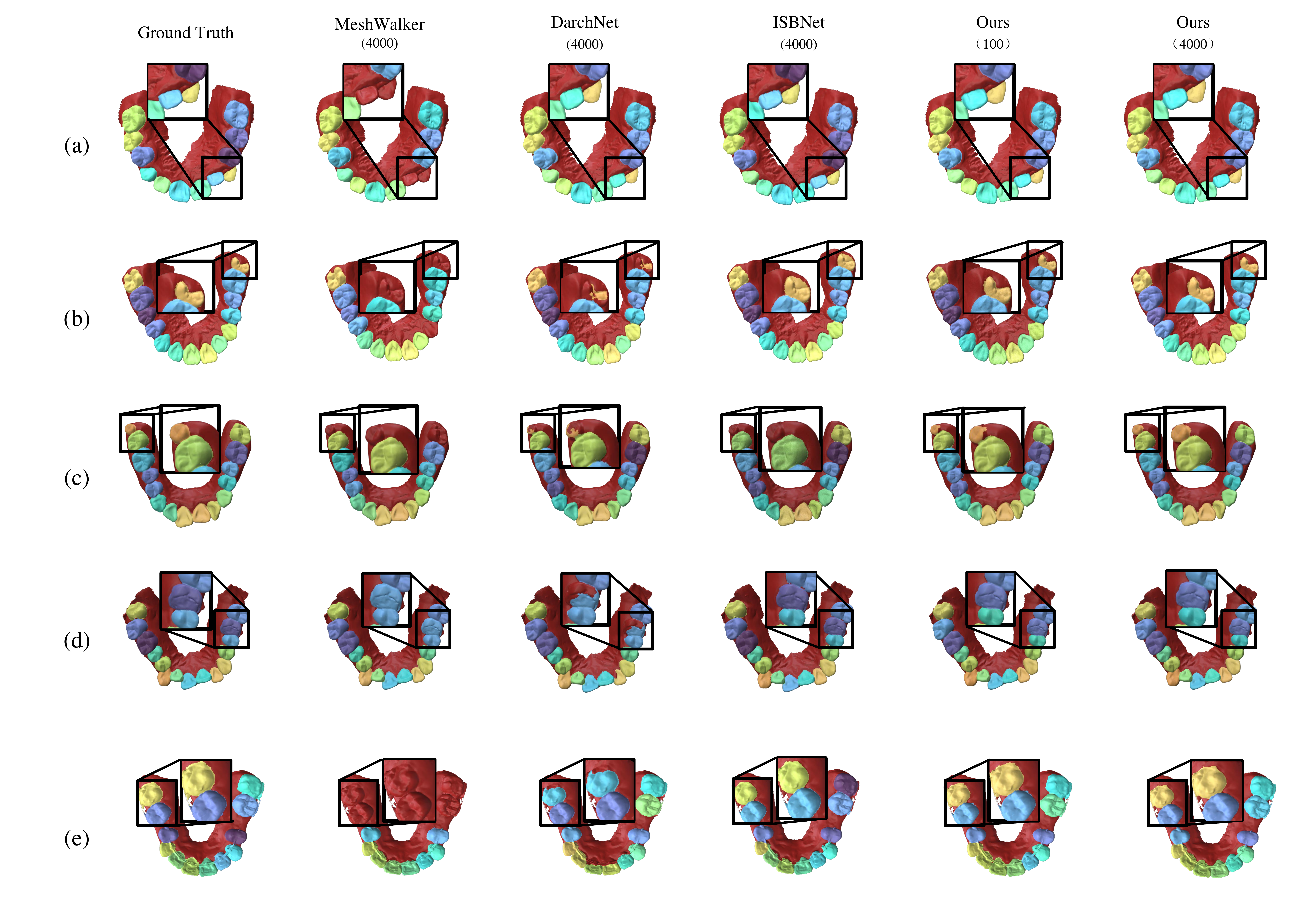}
  \caption{Segmentation results of SOTA methods. The numbers represent the sizes of training datasets. The segmentation results of five typical tooth meshes (a-e) are reported:  (a) normal teeth, (b) teeth with developmental deformities, (c) artificially implanted teeth, (d) malformed teeth that are packed together, and (e) teeth with braces (shape noise).    }
  \label{fig:cases}
\end{figure*}
\begin{table*}[htbp]
  \centering
  \caption{Segmentation accuracy of different methods (\%).}
    \begin{tabular}{lcccccccccc}
    \toprule
    \multicolumn{1}{c}{\multirow{2}[2]{*}{Dataset}} & \multicolumn{2}{c}{100 meshes} & \multicolumn{2}{c}{500 meshes} & \multicolumn{2}{c}{1,000 meshes} & \multicolumn{2}{c}{2,000 meshes} & \multicolumn{2}{c}{4,000 meshes} \\
    \cmidrule(r){2-3}\cmidrule(r){4-5}\cmidrule(r){6-7}\cmidrule(r){8-9}\cmidrule(r){10-11}
      & mIOU   & Dice  & mIOU   & Dice  & mIOU   & Dice  & mIOU   & Dice  & mIOU   & Dice \\
    \midrule
    MeshWalker~\cite{meshwalker} & 80.32 & 85.21 & 83.4  & 87.22 & 84.99 & 88.09 & 85.46 & 88.32 & 85.67 & 88.41 \\
    TSegNet~\cite{tsegnet} & 88.02 & 91.94 & 90.83 & 94.29 & 92.18 & 95.39 & 92.59 & 95.66 & 92.78 & 95.74 \\
    TeethGNN~\cite{teethgnn}  & 90.29 & 93.78 & 92.40  & 95.6  & 93.78 & 96.63 & 94.18 & 96.90  & 94.35 & 96.96 \\
    ISBNet~\cite{cai2019isbnet} & 90.12 & 93.62 & 92.34 & 95.58 & 93.74 & 96.60  & 94.16 & 96.89 & 94.34 & 96.95 \\
    DarchNet~\cite{darch}  & 89.64 & 93.11 & 92.56 & 95.61 & 94.21 & 96.91 & 94.67 & 97.19 & 94.87 & 97.28 \\
    MSFormer-Mask2Former-ResNet50 & 92.85 & 95.84 & 93.55 & 96.5  & 94.13 & 96.86 & 94.61 & 97.14 & 94.85 & 97.24 \\
    MSFormer-Mask2Former-SwinL & \textbf{94.56} & \textbf{97.12} & \textbf{95.12} & \textbf{97.33} & \textbf{95.58} & \textbf{97.49} & \textbf{95.97} & \textbf{97.61} & \textbf{96.18} & \textbf{97.68} \\
    MSFormer-EVA-Vit & \textbf{95.79} & \textbf{97.53} & \textbf{96.29} & \textbf{97.81} & \textbf{96.70}  & \textbf{97.99} & \textbf{97.08} & \textbf{98.12} & \textbf{97.29} & \textbf{98.19} \\
    \bottomrule
    \bottomrule
    \end{tabular}%
    \label{tab:acc}%
\end{table*}%
We conduct a comparative analysis between MSFormer and five prevalent tooth segmentation methods, including mesh-based methods (MeshWalker~\cite{meshwalker}), voxel-based methods (ISBNet~\cite{cai2019isbnet}), and point-cloud-based methods (TSegNet~\cite{tsegnet}, DarchNet~\cite{darch}, TeethGNN~\cite{teethgnn}). To ensure fair comparisons, we search for the optimal hyperparameters for each method.

\begin{itemize}
\item \textbf{MSFormer.} MSFormer employs well-trained multiview segmentation models to extract image features and generate segmentation results, which implies that MSFormer's performance may be enhanced through the use of more powerful multiview segmentation models. Thus, MSFormer is evaluated using three well-trained multiview pre-trained models: Mask2Former-ResNet50 (small)~\cite{m2former}, Mask2Former-SwinL (medium)~\cite{m2former}, and EVA-Vit (large)~\cite{eva}. Given that the EVA's pre-trained model mandates an input dimension of $1024 \times 1024$ pixels, we accordingly generate new multiview images. To facilitate compatibility with consumer-grade hardware during both training and testing phases, a low-rank adaptation technique~\cite{lora} is employed for the fine-tuning of pre-trained multiview models, specifically targeting the multiview perception and segmentation modules of MSFormer.
\item \textbf{DarchNet.} Following the recommendation provided in~\cite{darch}, we explore the number of graph convolution layers within $\{3,4,5\}$, and set the fixed crop size as 2048. DarchNet performs tooth mesh segmentation through a three-step process: first, it detects each tooth instance, followed by the extraction of a fixed-sized 3D patch for each instance, and finally, it segments the primary tooth from each 3D patch. However, this fixed crop size introduces inherent limitations. For instance, large teeth may be truncated if the fixed crop size is not large enough to cover all vertices of large teeth. Conversely, a sufficiently large crop size that includes both a large tooth and adjacent small teeth may result in the successful segmentation of the large tooth but the inadvertent omission of the small teeth, misclassified as background. Thus, DarchNet will suffer from incomplete large teeth or missing small teeth. 
\item \textbf{ISBNet.} ISBNet requires an initial phase of pre-training for semantic segmentation, followed by additional training geared toward instance segmentation.  Teeth and gums are categorized as distinct semantic entities, while individual teeth are viewed as separate instances within the broader tooth category. In accordance with the guidelines provided in ~\cite{cai2019isbnet}, we map the tooth mesh onto a cubic space defined by dimensions of $1 \times 1 \times 1$, set a ball query radius of 0.4, allocate a neighbor count of 32, and specify the layer size as 7.
\item \textbf{TeethGNN.} As stipulated by~\cite{teethgnn}, the downsampled mesh size is established at 15,000, the EdgeConv size is set to 5, and the number of nearest neighbors is determined to be 16. It is pivotal to note that TeethGNN employs varying hyperparameters to correct erroneous labels for upper and lower teeth, such as employing distinct fixed cut thresholds. Given that upper and lower tooth meshes may be intermixed in real-world applications, we adopt a mean-based approach for these differing parameters, resulting in a fixed cut threshold of 8.25 applicable to all meshes.
\item \textbf{TSegNet.} The search for an optimal point encoder length and sampling size encompasses the sets $\{3,4,5\}$ and $\{16000,20000,24000\}$, respectively. Analogous to DarchNet, TSegNet also employs a fixed crop size, leading to similar complications such as truncated large teeth or missing small teeth.  According to the suggestions in~\cite{tsegnet}, we set the crop size as 4096.
\item \textbf{MeshWalker.} We explore the optimal walk length and number of GRU layers from the sets $\{300,600,900\}$ and $\{3,6,9\}$, respectively. Given that mesh segmentation methods are not directly translatable to instance segmentation, we adapt MeshWalker to perform a closely related task: predicting instance boundaries to achieve instance segmentation. 
\end{itemize} 
\begin{table*}[htbp]
  \centering
  \caption{Number of parameters of different methods (in millions).}
    \begin{tabular}{lccccc}
    \toprule
    Parameters & Meshwalker & DarchNet & ISBNet & MSFormer-Mask2Former-SwinL & MSFormer-EVA-Vit \\
    \midrule
    Pre-trained parameters & 0     & 0     & 0     & 215.6 & 356.7 \\
    Non-pretrained parameters & 41.2  & 30.7  & 17.1  & 2.8   & 2.8 \\
    \bottomrule
    \bottomrule
    \end{tabular}%
  \label{tab:params}%
\end{table*}%
\begin{table*}[htbp]
  \centering
  \caption{Average segmentation time of different methods (in seconds).}
    \begin{tabular}{lccccc}
    \toprule
    Time & Meshwalker & DarchNet & ISBNet & MSFormer-Mask2Former-SwinL & MSFormer-EVA-Vit \\
    \midrule
    Non-deep-learning part & 2.95  & 0     & 0     & 2.56  & 3.77 \\
    Deep-learning part & 0.17  & 0.98  & 0.65  & 0.99  & 1.65 \\
    \bottomrule
    \bottomrule
    \end{tabular}%
  \label{tab:speed}%
\end{table*}%

Segmentation performance is assessed through well-established metrics, including mIOU (mean intersection over union) and Dice (dice coefficient)~\cite{darch}. All methods undergo training for more than 300 epochs, with only the best evaluation performances being reported.

\SubSection{Results}

\SubSubSection{Segmentation accuracy}
By training all models on datasets with different sizes, we obtain segmentation accuracy outlined in Tab.~\ref{tab:acc}. As the volume of training data increases, all methods exhibit enhanced performance; however, the gain gradually decreases and stabilizes. Utilizing large pre-trained vision models, MSFormer outperforms its segmentation results with just 100 training meshes compared with other methods trained on 4,000 meshes. This performance advantage is consistently maintained by MSFormer, exceeding 2\% with the expanding dataset. The segmentation results for SOTA methods, evaluated on five representative meshes, are depicted in Fig.~\ref{fig:cases}. The subsequent detailed analyses for each method are as follows:
\begin{itemize} 
  \item \textbf{MSFormer.} As indicated in Tab.~\ref{tab:acc}, the performance of MSFormer depends on the choice of pre-trained models. While utilizing Mask2Former-ResNet50, the pre-trained model leads to segmentation accuracy that is lacking to a certain extent. However, when employing larger vision models, such as Mask2Former-SwinL and EVA-Vit, the accuracy rapidly approaches the state-of-the-art performance (toward SOTA).  Fig.~\ref{fig:cases} demonstrates that MSFormer through the integration of priors and intrinsic knowledge embedded in pre-trained models excels in segmenting typical tooth meshes, even with 100 training data. The robustness of MSFormer is further validated by its superior handling of shape noise, as illustrated in Fig.~\ref{fig:cases}(e). Increasing the training dataset to 4,000 meshes equally enhances the method’s performance, particularly in refining segmentation boundaries.
  \item \textbf{DarchNet.} 
  DarchNet initially recognizes each tooth instance, followed by segmenting the individual teeth from cropped 3D patches. Unlike end-to-end models, two-stage models may accumulate errors at each processing stage. Lower detection accuracy in the initial stage can adversely affect the quality of subsequent segmentation, as evidenced by its lower performance with 100 meshes in Tab.~\ref{tab:acc}. Furthermore, DarchNet employs fixed-size cropping for each 3D patch, thereby potentially leading to incomplete cropping of large teeth or erroneously categorizing numerous small teeth as nonteeth during training. This drawback is visually represented in Fig.~\ref{fig:cases}(b,c), where the segmentation accuracy of the DarchNet for smaller teeth adjacent to larger teeth is noticeably inferior to that of MSFormer.
  \item \textbf{ISBNet.} Without large pre-trained models, existing methods struggle to attain satisfactory generalization on particular extreme cases, such as the closely packed teeth in Fig.~\ref{fig:cases}(d). ISBNet introduces innovative strategies to adeptly manage such situations, thereby delivering enhanced segmentation results. Nevertheless, ISBNet is not designed for tooth segmentation and ignores the prior shape knowledge relevant to teeth, such as tooth distribution~\cite{darch} and key shape features~\cite{teethgnn}. Consequently, even though ISBNet demonstrates superior performance in some extreme cases, it falls short when compared to SOTA methods explicitly designed for tooth segmentation, such as DarchNet.
  \item \textbf{TeethGNN.}  As expressed in~\cite{teethgnn}, TeethGNN employs distinct hyperparameters to optimize incorrect labels in the lower and upper teeth.  However, in practical applications where upper and lower teeth may be intermixed in the inputs of segmentation models, the adoption of a unified hyperparameter is an almost inevitable choice and compromises the segmentation results of the TeethGNN.  Consequently, TeethGNN lags behind other point-cloud-based approaches, such as DarchNet.
  \item \textbf{TSegNet.} Similar to DarchNet, TSegNet extracts teeth from fixed-size 3D patches and consequently incurs analogous challenges, including the incomplete segmentation of larger teeth and the neglect of smaller teeth (Fig.~\ref{fig:cases}(b,c)). Furthermore, TSegNet lacks additional measures to correct erroneous tooth centroids, rendering its performance inferior to DarchNet, as evidenced in Tab.~\ref{tab:acc}. 
  \item \textbf{MeshWalker}. This method employs instance boundary prediction for tooth segmentation tasks. Inaccuracies in predicting these boundaries can cause them to remain unclosed, leading to the omission of entire instances. As we cannot guarantee that all segmentation errors will occur outside these crucial boundary regions, even a few errors in segmentation boundaries may lead to a significant number of missing teeth. Fig.~\ref{fig:cases} supports our analysis, revealing an increase in missing instances, which leads to the lowest segmentation accuracy.
\end{itemize}

\begin{table*}[htbp]
  \centering
  \caption{Segmentation accuracy of different models in ablation study (\%).}
    \begin{tabular}{lcccccccccc}
    \toprule
    \multicolumn{1}{c}{\multirow{2}[4]{*}{Dataset}} & \multicolumn{2}{c}{100 meshes} & \multicolumn{2}{c}{500 meshes} & \multicolumn{2}{c}{1,000 meshes} & \multicolumn{2}{c}{2,000 meshes} & \multicolumn{2}{c}{4,000 meshes} \\
    \cmidrule(r){2-3}\cmidrule(r){4-5}\cmidrule(r){6-7}\cmidrule(r){8-9}\cmidrule(r){10-11}    
    & mIOU   & Dice  & mIOU   & Dice  & mIOU   & Dice  & mIOU   & Dice  & mIOU   & Dice \\
    \midrule
    MSFormer(Original) & \textbf{94.56} & \textbf{97.12} & \textbf{95.12} & \textbf{97.33} & \textbf{95.58} & \textbf{97.49} & \textbf{95.97} & \textbf{97.61} & \textbf{96.18} & \textbf{97.68} \\
    Model 1(No Skeleton) & 91.94 & 95.17 & 92.58 & 95.65 & 93.18 & 96.19 & 93.56 & 96.50  & 93.79 & 96.63 \\
    Model 2(Other skeletons) & 93.92 & 96.73 & 94.54 & 97.1  & 95.06 & 97.32 & 95.42 & 97.43 & 95.66 & 97.53 \\
    Model 3(No complement edge) & 93.11 & 96.15 & 93.72 & 96.59 & 94.28 & 96.94 & 94.64 & 97.16 & 94.82 & 97.23 \\
    Model 4(No contrastive model) & 92.61 & 95.68 & 93.24 & 96.28 & 93.81 & 97.67 & 94.23 & 96.92 & 94.42 & 97.05 \\
    Model 5(No contrastive loss) & 93.27 & 96.31 & 93.87 & 96.69 & 94.39 & 97.02 & 94.72 & 97.21 & 94.89 & 97.29 \\
    \bottomrule
    \bottomrule
    \end{tabular}
  \label{tab:abla}
\end{table*}
\begin{figure*}[ht!]
  \centering
  \includegraphics[width=0.8\textwidth,trim={4 4 4 4}, clip]{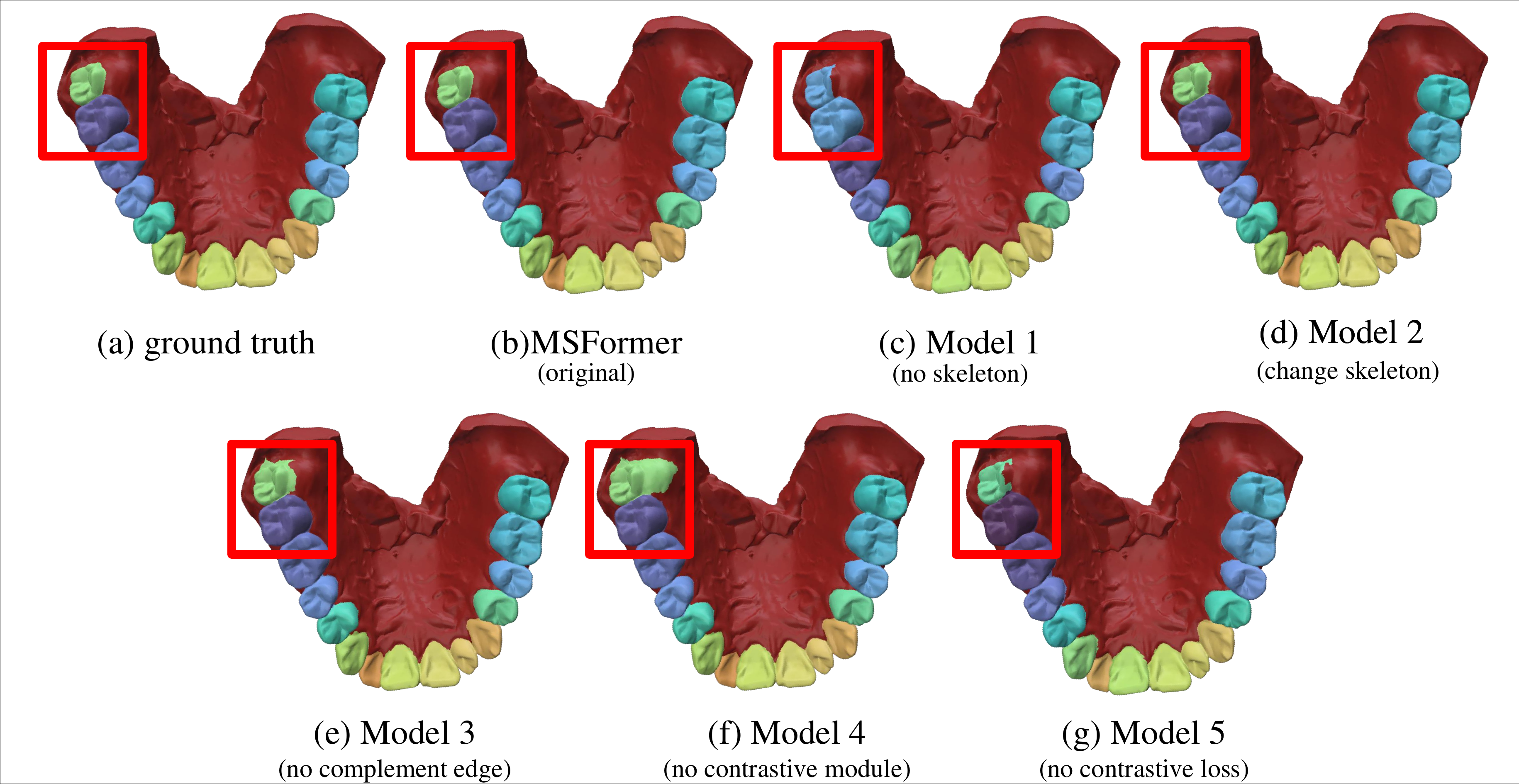}
  \caption{Segmentation results of different models in ablation study. }
  \label{fig:alb_cases}
\end{figure*}

\SubSubSection{Model complexity}
The parameter count for all methods is presented in Tab.~\ref{tab:params}.
MSFormer exhibits a higher number of parameters compared to other methods; however, a significant proportion of these parameters are derived from pre-trained models. These additional parameters contribute to MSFormer's superior performance by incorporating extensive prior knowledge. Specifically, when utilizing well-trained large vision models, MSFormer consistently outperforms other methods. Moreover, the quantity of non-pretrained parameters in MSFormer is substantially lower than in other approaches, facilitating effective training even with limited data. Consequently, the model complexity of MSFormer remains within acceptable ranges.
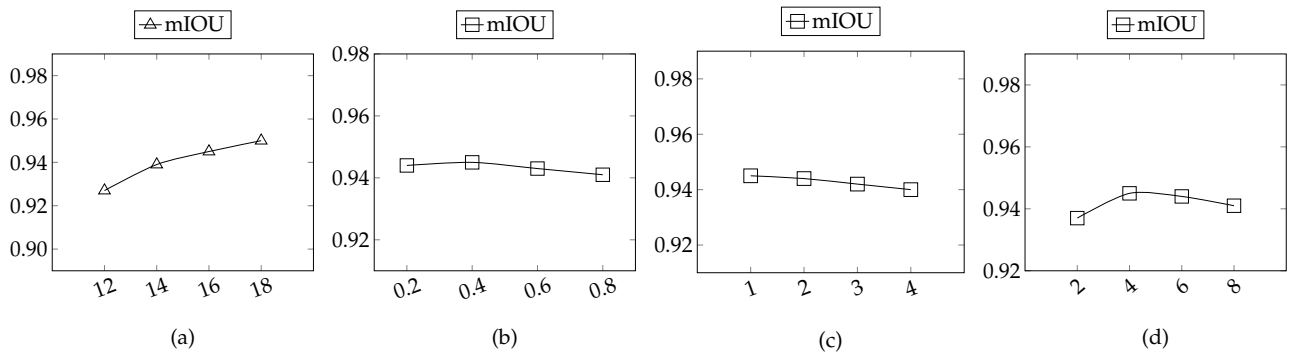
\begin{figure*}
  \begin{subfigure}{0.23\textwidth}
    \begin{minipage}[]{\textwidth}
      \centering 
      \resizebox{\linewidth}{!}{
        \begin{tikzpicture}
          \tikzstyle{every node}=[font=\fontsize{16}{16}\selectfont]
          \begin{axis}[
              xlabel= (a),
              % ylabel=$value$,
              xmin=10, xmax=20,
              ymin=89, ymax=99,
              xtick={12,14,16,18},
              x label style={at={(axis description cs:0.5,-0.15)},anchor=north},
              xticklabels={12,14,16,18}, 
              x tick label style={
                rotate=20
                            },
              ytick={90, 92, 94, 96, 98},
              yticklabels={0.90,0.92,0.94,0.96,0.98}, 
              legend style={at={(0.5,1.2)},anchor=north},
              legend columns=-1, 
                      ]
          \addplot[smooth,mark=triangle,black, width=3pt,mark size=5pt] plot coordinates {
            (12, 92.7)
            (14, 93.9)
            (16, 94.5)
            (18, 95.0)
          };
          \addlegendentry{mIOU}
          \end{axis}
          \end{tikzpicture}
        
      }
    \end{minipage}
  \end{subfigure}
  \begin{subfigure}{0.23\textwidth}
    \begin{minipage}[]{\textwidth}
      \centering 
      \resizebox{\linewidth}{!}{
        \begin{tikzpicture}
          \tikzstyle{every node}=[font=\fontsize{16}{16}\selectfont]
          \begin{axis}[
              xlabel=(b),
              x label style={at={(axis description  cs:0.5,-0.15)},anchor=north},
              % ylabel=$accuracy$,
              xmin=0.15, xmax=0.55,
              ymin=91, ymax=98,
              xtick={0.2, 0.3, 0.4, 0.5},
              xticklabels={0.2, 0.4, 0.6, 0.8}, 
              x tick label style={
                rotate=20
                            },  % <---
              ytick={92, 94, 96, 98 },
              yticklabels={0.92, 0.94,0.96,0.98},
              legend style={at={(0.5,1.2)},anchor=north},
              legend columns=-1, 
                      ]
          \addplot[smooth,mark=square,black,   width=3pt,mark size=5pt] plot coordinates {
            (0.2  ,94.4)
            (0.3  ,94.5)
            (0.4  ,94.3)
            (0.5  ,94.1)
          };
          \addlegendentry{mIOU}
          \end{axis}
          \end{tikzpicture}
      }
    \end{minipage}
  \end{subfigure}
  \begin{subfigure}{0.235\textwidth}
        \begin{minipage}[]{\textwidth}
          \centering 
          \resizebox{\linewidth}{!}{
            \begin{tikzpicture}
              \tikzstyle{every node}=[font=\fontsize{16}{16}\selectfont]
              \begin{axis}[
                xlabel=(c),
                xmin=0, xmax=5,
                x label style={at={(axis description  cs:0.5,-0.15)},anchor=north},
                ymin=91, ymax=99,
                xtick={1,2,3,4},
                xticklabels={1,2,3,4}, 
                x tick label style={
                  rotate=20
                              },  % <---
                ytick={92,94,96,98},
                yticklabels={0.92, 0.94,0.96,0.98},
                legend style={at={(0.5,1.2)},anchor=north},
                legend columns=-1, 
                          ]
              \addplot[smooth,mark=square,black, width=3pt,mark size=5pt] plot coordinates {
                (1,      94.5)
                (2,      94.4)
                (3,      94.2)
                (4,      94.0)
              };
              \addlegendentry{mIOU}
              \end{axis}
              \end{tikzpicture}
          }
    \end{minipage}
  \end{subfigure}
  \begin{subfigure}{0.23\textwidth}
      \begin{minipage}[]{\textwidth}
        \centering 
        \resizebox{\linewidth}{!}{
          \begin{tikzpicture}
            \tikzstyle{every node}=[font=\fontsize{16}{16}\selectfont]
            \begin{axis}[
                xlabel=(d),
                x label style={at={(axis description  cs:0.5,-0.15)},anchor=north},
                % ylabel=$value$,
                xmin=0, xmax=10,
                ymin=92, ymax=99,
                xtick={2, 4, 6, 8}, 
                xticklabels={2, 4, 6, 8},
                x tick label style={
                  rotate=20
                              },   % <---
                ytick={92,94,96, 98},
                yticklabels={0.92,0.94,0.96,0.98},
                x tick label style={
                  rotate=20
                              },
                legend style={at={(0.5,1.2)},anchor=north},
                legend columns=-1, 
                        ]
            
            \addplot[smooth,color=black,mark=square, width=3pt,mark size=5pt]
                plot coordinates {
                  (2, 93.7)
                  (4, 94.5)
                  (6, 94.4)
                  (8, 94.1)
                };
            \addlegendentry{mIOU}
            \end{axis}
            \end{tikzpicture}
  
        }
      \end{minipage}
    \end{subfigure}
    \caption{(a) Segmentation performance under various numbers $m$ of images. 
    (b)  Segmentation performance under various pooling rates $r$.
    (c) Segmentationn performance under various numbers of pooling operation.
    (d) Segmentation performance under different $k$ for contrastive loss.}
    \label{fig:hyper}
  \end{figure*}

\SubSubSection{Time complexity}
We evaluated the average computational time across 8,000 tooth meshes for SOTA methods, the results of which are documented in Tab.~\ref{tab:speed}. The computational time can be bifurcated into non-deep-learning-based and deep-learning-based parts. The former does not exclusively rely on the computational device, prompting us to employ ten parallel processes for this part and a singular process for the deep-learning-based part. Tab.~\ref{tab:speed} reveals that the non-deep-learning-based part does necessitate additional computational time compared to other methods; however, the increase remains within an acceptable range, approximately 2.18 seconds. In the case of the deep-learning-based part, MSFormer, when integrated with large vision models, also incurs a longer computational time. Nevertheless, relative to the average time, the average increment in computational time for MSFormer is a mere 0.72 seconds.

\SubSection{Ablation study}
\label{sec:abla}
MSFormer achieves the best segmentation results in tooth segmentation, attributing its exceptional performance to three pivotal factors:  external skeleton data, a lightweight skeletonNet, and a lightweight contrastive learning module. Among various versions of MSFormer, MSFormer-Mask2Former-SwinL offers an optimal trade-off between performance and computational speed, rendering it highly applicable across diverse settings. To elucidate the individual contributions of these three factors, we conducted an ablation study on MSFormer-Mask2Former-SwinL.

\textbf{Benefits of skeleton data. } Initially, to evaluate the impact of skeleton data, Model 1 degrades to a pure multiview-based method by removing all skeleton-related modules. As elaborated in Sec.~\ref{sec:intro}, multiview-based methods may introduce a biased and incomplete perception of shapes, thereby adversely affecting segmentation results. Fig.~\ref{fig:alb_cases} corroborates that exclusive reliance on multiview perception induces increased errors, such as omitting or erroneously merging teeth during segmentation. Conversely, the integration of skeleton data facilitates a 2D–3D joint perception, yielding more accurate segmentation results. Additionally, skeletons generated via various algorithms might produce different results. We employ other popular algorithms, such as hierarchical clustering (Model 2), for skeleton generation. Nonetheless, as indicated by the accuracy metrics in Tab.~\ref{tab:abla}, spectral clustering (employed in MSFormer) appears to outperform hierarchical clustering. A plausible explanation is that the spectral clustering algorithms are specifically tailored for mesh segmentation, thereby offering superior pre-segmentation~\cite{spectral}. 

\textbf{Benefits of SkeletonNet.} 
An essential contribution of SkeletonNet is its ability to preserve connectivity during the downsampling process, accomplished through edge complementation.  As illustrated in Fig.~\ref{fig:skeletonNet}(a), if the complemented edge is eliminated, certain topological information will inevitably be lost during downsampling. Therefore, Model 3 is constructed by removing these complemented edges to assess the impact of losing topological information. Fig.~\ref{fig:alb_cases} illustrates that the segmentation results of Model 3 exhibit a greater number of errors compared with the original MSFormer. Tab.~\ref{tab:abla} quantitatively provides this decrease in accuracy.

\textbf{Advantages of the contrastive learning module.}  We construct Model 4 by substituting the contrastive learning module with feature concatenation to replace the contrastive learning module and investigate the gains of the contrastive learning module. Specifically, the average features of skeleton nodes will be  concatenated  to the  feature of one image patch if the faces corresponding to these skeleton nodes are projected to this image patches. In addition, we remove contrastive learning loss in Model 5 to gauge its significance. Model 4 uses an incomplete relationship between skeleton nodes and image patches to achieve feature fusion, which may cause the failure of feature fusion. Thus, Model 4 may degrade to multiview-based methods, suffering from biased 3D perception and more errors, as shown in Fig.~\ref{fig:alb_cases}. Similarly, without our contrastive learning loss, MSFormer might be unable to capture a correct and comprehensive relationship between skeleton nodes and image patches under limited data source. As shown in Tab.~\ref{tab:abla}, neither Model 4 nor Model 5 can achieve competitive segmentation results.

\SubSection{Hyperparameter analysis}
\label{sec:hyper}
We analyze the sensitivity of MSFormer to critical hyperparameters, including the number $m$ of multiview images, the pooling rate $r$ of SkeletonNet,  the number of pooling operations, and the $k$ value of our contrastive loss. This analysis aims to offer insights for developing a more robust method. MSFormer-Mask2Former-SwinL is trained with various hyperparameters utilizing 100 meshes, revealing the sensitivity to these hyperparameters.

Furthermore, we examine the influence of different numbers of images by setting the number of multiview images $m$ as $\{12, 14, 16, 18\}$. As depicted in Fig.~\ref{fig:hyper}(a), increasing the number of images enhances accuracy. However, upon reaching $m = 18$, consumer-grade hardware (e.g., RTX 3090) cannot support the training progress. Consequently, we conducted the training on a more advanced machine (Tesla A100 40G and AMD 5950X). Given that 18 images do not produce a significant improvement, we set $m = 16$.

A high pooling rate $r$ results in excessive downsampling of skeletons, whereas a low rate requires additional convolutions and an increased parameter count. Both extremes can adversely affect segmentation results. To identify an optimal $r$, we experiment with values in the range  $\{0.2, 0.4, 0.6, 0.8\}$. According to the findings presented in Fig.~\ref{fig:hyper}(b), an $r$ of 0.4 achieves the best performance.

Moreover, we investigate the influence of SkeletonNet’s size by altering the number of pooling operations in SkeletonNet, exploring configurations within the set $\{1, 2, 3, 4\}$. Contrary to expectations, an enlarged SkeletonNet does not enhance segmentation accuracy. With only 100 training meshes available, a large SkeletonNet may lack sufficient data for effective training, thereby compromising segmentation results. Based on the data presented in Fig.~\ref{fig:hyper}(c), we elect to set the number of pooling operations to 1.

Additionally, we discuss the potential effects of different values of $k$ by setting it to $\{2, 4, 6, 8\}$. An excessively large $k$ can yield inaccurate relationships between skeleton nodes and image patches. Conversely, we risk an incomplete relationship if k is too small. According to Fig.~\ref{fig:hyper}(d), the optimal segmentation performance is attained when $k = 4$.

\SubSection{Discussion}

\textbf{Advantages of MSFormer:} MSFormer achieves 2D–3D joint perception by incorporating two lightweight modules into pre-trained multiview-based models. This strategy not only reduces the required training data of a 2D–3D joint model but also delivers superior segmentation performance with just 100 meshes. One direct benefit of MSFormer is its lower data preparation costs compared to existing methods. Additionally, as fewer tooth meshes are needed, ethical concerns associated with collecting personal biological data can be alleviated. Furthermore, MSFormer exhibits favorable scalability. By employing a plug-and-play approach, these added lightweight modules could enhance other multiview pre-trained models with a more powerful performance in future research endeavors.

\textbf{Limitation of MSFormer:} First, MSFormer operates based on the statistical assumption that complex hollow structures on tooth meshes are not prevalent. If tooth meshes contain numerous intricate hollow structures, it is challenging for 16 multiview images to adequately capture all instances, thereby leading to worse segmentation results. However, according to Tab.~\ref{tab:viewnumber} in the appendix, such intricate hollow structures in tooth meshes are relatively uncommon. Second, MSFormer depends on convolutions to extract 3D perception, meaning that skeletons must maintain robust connectivity. If non-connected meshes emerge due to irregularities in data collection, the effectiveness of MSFormer may be reduced, causing it to function more as a multiview-based segmentation method.

%% --------------------------------

\Section{Conclusion}
This paper introduces MSFormer, a 2D–3D joint method, for tooth instance segmentation. By incorporating two lightweight and meticulously designed modules to extract and replenish 3D perception, MSFormer overcomes incomplete perception obtained from 2D images and achieves superior segmentation results, even with limited training data. Experimental results demonstrate that MSFormer combined with large vision models achieves an improvement of over a 2\% accuracy compared to existing tooth segmentation methods, regardless of the quantity of training data. Although MSFormer currently exhibits long computational times for segmentation tasks and limited capability in handling non-connected tooth meshes, it is pertinent to acknowledge that future refinements may mitigate these limitations. One possible approach for reducing computational time is to employ model distillation techniques; additional cross-modal data, such as volumetric data, may help efficiently address the irregular cases of non-connected tooth meshes.

%% -----------------------------

% \IEEEpeerreviewmaketitle
% \input{sections/acknowledgments}

\ifCLASSOPTIONcaptionsoff
  \newpage
\fi
\bibliographystyle{IEEEtran}
\bibliography{IEEEabrv,biblio}

\clearpage

\Section{Appendix}
\label{sec:sup}
 
% \begin{figure*}[ht!]
%   \centering
%   \includegraphics[width=0.8\textwidth,clip]{fig/poseture.jpg}
%   \caption{Align the postures of different tooth meshes. }
%   \label{fig:find_axis}
% \end{figure*}
\begin{figure*}[ht!]
  \centering
  \includegraphics[width=0.8\columnwidth,trim={2 2 2 2},clip]{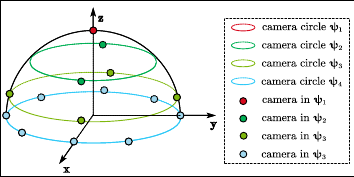}
  \caption{A half of spherical camera space.  In this example, the multiview camera setting is divided into four camera circles, i.e., $\varphi = [\varphi_1, \varphi_2, \varphi_3, \varphi_4] = [1,2,4,8]$.  Since camera circles are evenly distributed into this camera space, the angle between the line connecting the camera position to the coordinate origin and the xy plane is $[\pi/2, \pi/3, \pi/6, 0]$ for cameras in four camera circles, respectively. Cameras in each circle are distributed uniformly along the circumference.  }
  \label{fig:cameraspace}
\end{figure*}
\begin{figure*}[ht!]
  \centering
  \includegraphics[width=0.8\textwidth,clip]{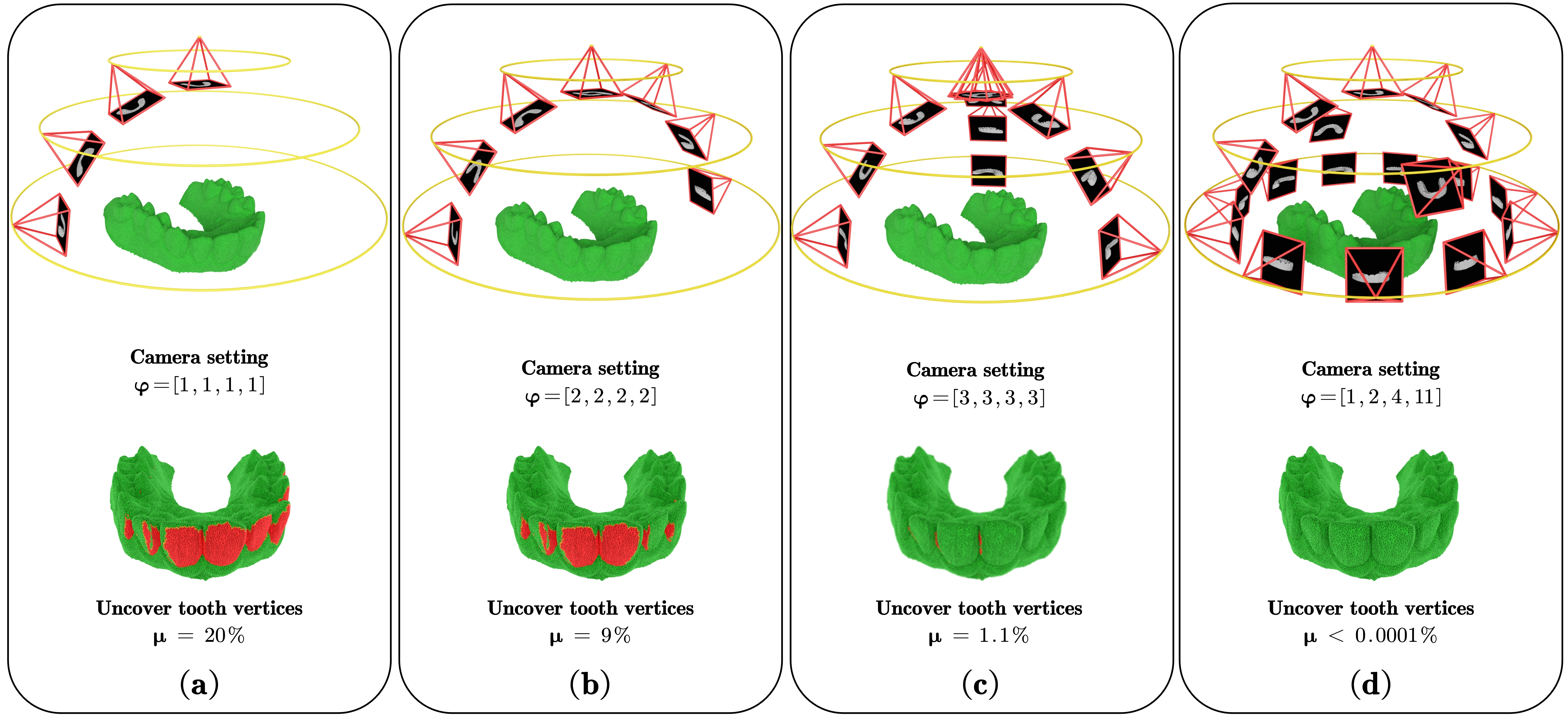}
  \caption{Percentage $\mu$  of covered 
  vertices across  all instances in different multiview camera settings, i.e., ratio of visible tooth vertices in multiview images to all vertices of tooth instances. }
  \label{fig:converage}
\end{figure*}
\begin{table*}[htbp]
  \centering
  \caption{The percentage $\mu$  of covered vertices across  all instances in different view sizes. We search for the optimal camera setting in the spherical camera space and calculate the percentage of covered vertices across all instances. Specifically, if any vertex distance from the camera center is greater than or equal to the z-buffer value of the projection position, the vertice is considered visible. (\%) }
  \begin{tabular}{lcccccccccccc}
    \toprule
    \multirow{2}[0]{*}{Covered tooth vertices ($\mu$)} & \multicolumn{3}{c}{ 12 views} & \multicolumn{3}{c}{14 views} & \multicolumn{3}{c}{16 views} & \multicolumn{3}{c}{18 views} \\
    \cmidrule(r){2-4}\cmidrule(r){5-7}\cmidrule(r){8-10}\cmidrule(r){11-13}
          & min   & mean  & max   & min   & mean  & max   & min   & mean  & max   & min   & mean  & max \\
    \midrule
    Optimal camera setting & 91.65 & 96.99 & 99.98 & 95.12 & 99.13 & 100   & 96.41 & 99.91 & 100   & 96.75 & 99.93 & 100 \\
    \bottomrule
    \bottomrule
    \end{tabular}%
  \label{tab:viewnumber}%
\end{table*}%

\SubSection{Optimal multiview camera setting}
\label{sec:multiviewsetting}
Determining the optimal configuration for multiview cameras, specifically the number $m$ of multiview images, constitutes a formidable challenge. As illustrated in Fig.~\ref{fig:converage}(a-d), increasing the number of multiview images expands the coverage of tooth instances. Ideally, an infinite array of multiview images would offer comprehensive coverage of every instance. However, limitations in computational resources, such as GPU memory, necessitate a more prudent selection of $m$.

The optimal scenario entails identifying the minimum number of multiview images, denoted as $m_{min}$, which ensures coverage of most vertices for each tooth instance (over 99.9\%), all while adhering to computational constraints. If tooth instances possess complex hollow structures, an extensive array of multiview images may be obligatory to achieve full coverage. In such instances, $m_{min}$ may need to be significantly large, thereby rendering multiview-based segmentation approaches impracticable. Fortunately, tooth meshes generally lack such complex hollow structures, particularly in tooth segmentation. In cases featuring non-hollow structures and well-aligned postures, a limited number of images suffice to cover the vast majority of each tooth instance's vertices. This observation has been corroborated by data presented in Tab.~\ref{tab:viewnumber}. Consequently, within the context of the prevalent spherical camera space, the procedure for identifying $m_{min}$ can be summarized accordingly.

 For tooth meshes exhibiting aligned postures, all tooth instances predominantly reside within one half of the spherical camera space, as shown by Fig.~\ref{fig:cameraspace}. Consequently, our search for the optimal camera setting is confined to this half, as illustrated in Fig.~\ref{fig:converage}. We divide a multiview camera setting $\varphi$ into $s$ uniformly distributed camera circles $[\varphi_1, ...,\varphi_i,..., \varphi_s]$ within this half of the spherical camera space, where $\varphi_i$ means the number of cameras in the $i$-th camera circle, and the sum of $(\varphi_1,..., \varphi_s)$ represents the number $m$ of multiview images. Given the possible number $m \in \{12, 14, 16, 18\}$ of images and the number $s \in \{2, 3, 4, 5\}$ of camera circles, we undertake an exhaustive exploration of all feasible multiview camera settings $\varphi = [\varphi_1, ...,\varphi_s]$ and calculate the optimal camera setting that can cover the most vertices of each tooth instance with $m_{min}$. Since cameras within each circle are uniformly distributed along the circumference, further adjustments to their positions become superfluous, as shown in Fig.~\ref{fig:cameraspace}. Extending the search range for $m$ may be unnecessary, as consumer-grade hardware may lack the capacity to support more than 18 views, particularly when employing large vision models. Statistical analyses on 8000 meshes reveal that the camera setting $m_{min}=16$, $s=4$, $\varphi = [1, 2, 4, 9]$ comprehensively covers the vertices of each tooth instance—exceeding 99.9\% coverage on average—as listed in  Tab.~\ref{tab:viewnumber}.

 Thus, we find the optimal camera setting for multiview-based tooth segmentation, the performance of which has been substantiated through extensive testing on a comprehensive real-world dataset. This optimal number of multiview images does not surpass computational limitations and achieves coverage of over 99.9\% of the vertices for each tooth instance on average, thereby indirectly affirming the viability of multiview-based tooth segmentation.

\end{document}